\documentclass[11pt]{article}

\usepackage[final]{acl}

\usepackage{times}
\usepackage{latexsym}
\usepackage[T1]{fontenc}
\usepackage[utf8]{inputenc}
\usepackage{microtype}
\usepackage{inconsolata}
\usepackage{amsmath}
\usepackage{amssymb}
\usepackage{amsthm}
\usepackage{graphicx}
\usepackage{mathtools}
\usepackage{listings}
\usepackage{tikz}
\usepackage{siunitx}
\usepackage{tipa}
\usepackage{float}
\usepackage{bbm}
\usepackage{breqn}
\usepackage{utils}
\usepackage{tikz-dependency}
\usepackage{nicefrac}
\usepackage{booktabs}

\usepackage{caption}
\usepackage{subcaption}
\usepackage{linguex}

\title{(V)LMs generalize beyond surface co-occurrence:\\Evidence from cross-modal number agreement}

\author{Zach Studdiford \\
  Department of Psychology \\
  University of Wisconsin-Madison \\
  \texttt{studdiford@wisc.edu} \\\And
  Kanishka Misra \\
  Department of Linguistics \\
  The University of Texas at Austin \\
  \texttt{kmisra@utexas.edu} \\}

\newcommand{\wug}{\texttt{[wug]}}
\newcommand{\wugs}{\texttt{[wugs]}}

\begin{document}
\maketitle
\begin{abstract}

Language models learn about grammatical number primarily from co-occurrence, and show frequency effects as a result---sometimes taken to indicate that they do not learn abstract ``rules'', and are instead dependent on specific lexical items. Testing generalization with text stimuli alone cannot settle this debate, since distributional cues (is/are, this/these) easily give number away. We instead use cross-modal generalization as a tool to investigate abstractions in LMs that can also accept visual inputs (VLMs), restricting the evidence that diagnoses number to an extra-linguistic modality. We teach VLMs pairs of new nouns by adding new embeddings and only updating them during learning, comparing conditions where number is diagnosed by visual cues alone against ones where it is disambiguated by text. Across behavior, representational dynamics, and causal mechanisms, we find non-trivial evidence for cross-modal generalization across both exposure conditions, and that linguistic vs. extra-linguistic cue conditions are treated in similar ways in the internal mechanisms of the model. This suggests that statistical learners like VLMs can generalize beyond surface-level co-occurrence and show genuine abstraction-compatible behavior.

% generalization is above chance in both conditions; the novel embeddings move towards regions inhabited by real nouns with the same number feature; and mechanisms discovered before the novel words were acquired remain causally efficacious for them, with no effect of cue condition. Statistical learners like VLMs can therefore go beyond surface-level co-occurrence, and show genuine abstraction-compatible behavior.
\end{abstract}

\section{Introduction}
\label{sec:intro}

% jeff and dave are said to be jeff elman and dave rumelhart respectively.
Jeff and Dave are walking in an art museum where they encounter a painting of a single, cute looking animal called ``\textit{snarpus}''. Next to it, Jeff sees a painting of multiple such animals, with the label: ``\textit{snarpi}''. Even though Jeff has never heard of \textit{snarpus} or \textit{snarpi} before today, he says to Dave that ``\underline{these} \textit{snarpi} \underline{are} so cute!'', using the demonstrative ``these'' as opposed to ``this'', and the plural form \textit{are} as opposed to \textit{is}, owing to the fact that \textit{snarpi} was used in reference to more than one of the animals, and is therefore the plural of \textit{snarpus}. The made up story that we have just described is rather mundane, but it highlights our ability to abstract away our knowledge of \textit{grammatical number} in a manner that goes beyond specific modalities. 

\begin{figure}[!t]
    \centering
    \includegraphics[width=0.8\linewidth]{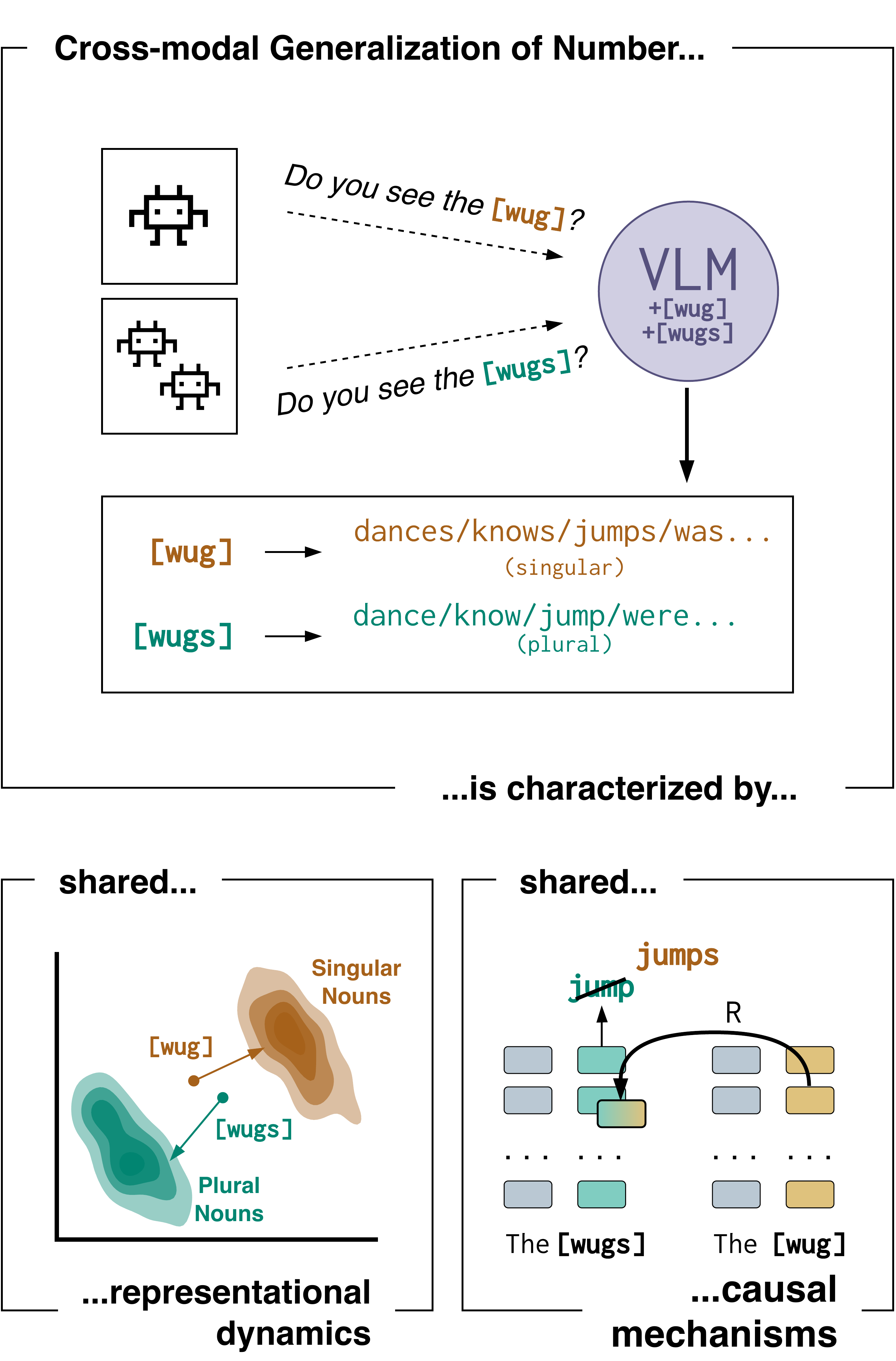}
    \caption{We use cross-modal generalization as a tool to investigate abstractions in VLMs. Our case study focuses on number-agreement, where we teach VLMs new words (here, \texttt{[wug]} and \texttt{[wugs]}) that share the same textual exposure but are differentiated (in terms of number) visually. We show that VLMs can successfully generalize from this learning setup to number-agreement judgments, that generalization is characterized by movement of the new words' embeddings towards number-based regions in the model's representational space, and is also accommodated in causal mechanisms that were discovered \textit{before} the VLMs learned the new words.}
    \vspace{-1em}
    % \caption{Given an interpretability method that localizes the mechanism for a phenomenon in a given model, we test if the mechanism can account for model behavior when the knowledge critical to the phenomenon comes from a different modality. We apply this method to \textit{number agreement} when the model learns about two nouns (\texttt{[wug]} and \texttt{[wugs]}) and their number feature through visual inputs, without access to orthography. \km{caption and annotations need to change.}
    % Given an LM and an interpretability method for a phenomenon (in this paper, \textit{number agreement}), we test for generalization when the model has to operate over new tokens (here, \texttt{[wug] and \texttt{[wugs]}}), whose phenomenon-specific knowledge (here, the nouns' \textit{number}) is acquired by the model through a different modality.
    % }
    \label{fig:fig1}
    % \vspace{-1.5em}
\end{figure}

% The goal of this paper is to show how such cross-modal generalizations can be a useful way to diagnose abstractions in neural network language models that also learn to accept inputs from a different modality. Specifically, 
Neural network language models (LMs) learn about number agreement from co-occurrence---dog occurs more often with \textit{is, barks,} etc. and dogs with \textit{are, bark, }etc. \citep{wei-etal-2021-frequency, hobbs-mccoy-2026-collocational}. This facet of models can (and has) been interpreted broadly in two different ways. Some suggest that by only learning from co-occurrence, LMs are prone to show disproportionately successful number agreement behavior on words for which they have seen a good deal of evidence \citep{wei-etal-2021-frequency, lasri-etal-2022-bert, wilson-etal-2023-abstract}. This finding is often taken to mean that LMs do not necessarily learn abstract ``rules'', and instead are highly dependent on specific lexical items \citep{lasri-etal-2022-bert, wilson-etal-2023-abstract, oba-etal-2024-language}. Others have instead shown that co-occurrence between specific nouns and verbs (e.g., ($\langle$\textit{cat, meows}$\rangle$, $\langle$\textit{lions, roar}$\rangle$) is in fact an important cue for learners (LMs and Humans alike) to acquire abstract generalization in making number agreement predictions in the first place \citep{hobbs-mccoy-2026-collocational}. Based on this latter viewpoint, abstractions are seen more as an emergent phenomenon that arise \textit{via} accumulation of evidence from individual exemplars, rather than something that has to be built explicitly into a learner \citep{ambridge2020against, misra-and-kim-2023-catabs, jian-manning-2026-humans,Dubova_Sloman_2026}.\footnote{Our aim is not to dispute the presence of clear item-specific frequency effects we see in models \citep{mccoy2024embers}---frequency effects are inevitable in any system that performs statistical learning, and are thoroughly prevalent in humans as well \citep{lupyan2013difficulties, ambridge2015ubiquity, lampinen2024language, studdiford2026reasoning}}

A historically productive method to investigate abstractions in children has been to conduct novel word learning trials \citep{berko1958child, hohle2004functional}. In a nutshell, these experiments show children new words using a combination of pictures and/or sentence utterances, and then test how they respond to the usage of the word in a different context. This has also been done for LMs in their learning of nouns and verbs \citep{kim-smolensky-2021-testing, wilson-etal-2023-abstract, misra2024generating}, but in the case of number, the exposure can easily give away information via orthography or by distributional cues (verbs like \textit{is/are}, demonstrative like \textit{these/this}, etc.). To what extent does a model encode abstract knowledge that cannot be explained merely by co-occurrence alone? 

To answer this question, we turn to novel word learning in variants of LMs that can process inputs from extra-linguistic modalities, where we can restrict diagnostic information about the target abstraction (number) for the novel words to come from a non-linguistic modality. We then measure the extent to which models are able to learn about the number agreement for these novel words to conclude about the strength of this abstraction. \textbf{That is, we use cross-modal generalization} \citep{xu-etal-2026-cross} \textbf{as a tool to explore abstractions in (V)LMs.}
% In our case, we use vision as our extra-linguistic modality, and study Vision-Language Models (VLMs). That is, we use cross-modal generalization to explore the strength of the abstractions learned by (V)LMs.
We do so by providing evidence from tests of model behavior, internal representational dynamics, and a battery of causal interpretability techniques. 

Specifically, we teach the model pairs of new nouns that refer to objects introduced in images, by using them in textual descriptions that are uninformative with respect to number. This way surface-form co-occurrence alone cannot diagnose the novel nouns' number. We do this by inserting new embeddings for these novel words, which then prevents the model from making conclusions using the surface form (e.g., using -s to conclude that it is plural) \footnote{We note that the VLMs we evaluate also do not have access to orthography for any sampled real nouns---i.e., both singular and plural nouns are tokenized into single tokens.}. During learning, we only update the embeddings of the novel words, so that we can analyze and conclude about how the learned abstractions (insofar as they exist) in their pre-existing representations pressure the learning of new information. We compare this to a setting where there is no image and the model instead learns from textual cues, where the linguistic cues clearly diagnose number information. We then measure generalization by performing standard minimal pair analysis on English number-agreement stimuli that are disjoint from those in training. We observe that performance on nouns learned from both visual as well as linguistic cues is significantly above chance, and degrades gracefully as the
number of intervening attractor nouns increases.
This suggests that behaviorally, models are able to demonstrate non-trivial cross-modal generalization of number.

Next, we characterize what underlies this behavior by investigating the dynamics of the embeddings of novel words during learning. We find that learning in both cue-conditions (vision vs. language) corresponds to similar representational behavior, where novel nouns move to regions in embedding space that are inhabited by real nouns that organize themselves in terms of their number features. That is, the embedding for the novel word that is intended to be treated as singular moves towards a region of known singular nouns (e.g., dog, cat, leopard, etc.) and its plural counterpart moves towards a region of known plural nouns (e.g., dogs, cats, leopards, etc.). This suggests that these low dimensional regions sensitive to grammatical number act as basins into which novel words tend to move towards during learning. 

Finally, we test if model mechanisms discovered using interpretability techniques \textit{before} novel words were acquired can readily accommodate knowledge of novel words. Using four different methods, we find this to be true---the causal efficacy of all methods tracked the results from running these methods on sentences with real nouns. We additionally found no difference in the methods' effectiveness for nouns acquired from language versus vision, suggesting that models could be integrating both types of nouns into their existing number-agreement mechanism in a similar manner. 

% Overall, our findings suggest that...
Overall, our findings suggest that even though VLMs learn linguistic features primarily through co-occurrence, and therefore display item-specific frequency effects, this does not necessarily indicate an absence of abstraction-compatible representations \citep{misra-and-kim-2023-catabs}. In fact, the representations that have emerged as a result of LM and VLM training can, in principle, flexibly accommodate linguistic information for tokens acquired primarily through extra-linguistic modalities. This suggests that statistical learners like VLMs can go beyond simple surface-level co-occurrence and integrate linguistic cues even from evidence completely devoid of explicit linguistic signal.

\section{Related Work}
% \paragraph{Number Agreement in Neural Networks} 
Number agreement has been perhaps the oldest probe for syntactic competence in neural network models of language \citep{elman1991distributed}. In more modern instantiations, it has been evaluated via minimal pair judgments \citep{linzen2016assessing, hu-etal-2020-systematic, mueller-etal-2020-cross} as well as through representational analyses \citep{lakretz-etal-2019-emergence, finlayson-etal-2021-causal, lasri-etal-2022-probing, arora-etal-2024-causalgym, marks2025sparse}. 
% It has also been the poster child phenomenon for many newer causal interpretability methods \citep{arora-etal-2024-causalgym, marks2025sparse, arora2026language}. 
It is also a common case-study for investigating language model generalization, and in particular has been shown to be affected by frequency effects \citep{wei-etal-2021-frequency, lasri-etal-2022-bert}, where models succeed on noun-verb pairs that are sufficiently frequent. 
Our work builds on the historical precedence of number-agreement by treating it as the target abstraction that we investigate in VLMs. We specifically test if models can demonstrate this knowledge in a manner that cannot be explained by the aforementioned frequency effects. We do this by conducting novel word learning studies where the knowledge of a noun's number is only diagnosed by extra-linguistic evidence, and comparing it to a case where it is diagnosed by explicit linguistic context. This allows us to test if the representations that result from (V)LM training can demonstrate abstraction-compatible behavior, even though they are primarily acquired from co-occurrence.

The method that we use to conduct novel word learning (\Cref{sec:embed_train}) also traces back to early connectionist approaches \citep{rumelhart1993learning, rogers2004semantic}, where new concept and property nodes were added to a trained network that predicted concept-property associations, trained on a set of inputs by only updating the new node representations, and then tested for inductive generalization. Since then, this method has been used to conduct word learning experiments \citep{lampinen2017one}, test for category learning \citep{kim-smolensky-2021-testing, misra-and-kim-2023-catabs}, structural alternations \citep{wilson-etal-2023-abstract, misra2024generating}, and neologisms \citep{hewitt2025we}. 
We extend this method by adding an additional modality during learning (vision), and conduct further analyses of how knowledge acquired through this method is integrated into the model by investigating representational dynamics of novel word embeddings, as well as how information about their number is accounted for by interpretability methods that discover mechanisms \textit{before} the novel words were acquired by the model. 
\section{General Methods}
% \label{sec:nwl}
\label{sec:methods}

In this section, we describe our method for learning novel embeddings in VLMs,
% (shown in Figure \ref{fig:full-methods-fig})
as well as evaluation stimuli, and the models studied. While we focus on number-agreement in this work, our training paradigm can in principle be extended to assess cross-modal generalization for a variety of linguistic abstractions.
% describe our approach for few-shot learning of singular and plural noun embeddings in VLMs. While we focus on VLM generalization for number agreement, our training paradigm can in principle be extended to assess cross-modal generalization for a wide variety of linguistic abstractions.

\begin{figure*}[t]
    \centering
    \footnotesize
    % ---------------- Vision: training ----------------
    \begin{subfigure}[t]{0.48\textwidth}
        \centering
        \begin{tabular}{@{}cc@{}}
            \includegraphics[height=0.95in]{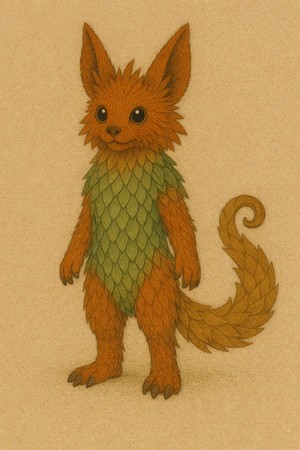} &
            \includegraphics[height=0.95in]{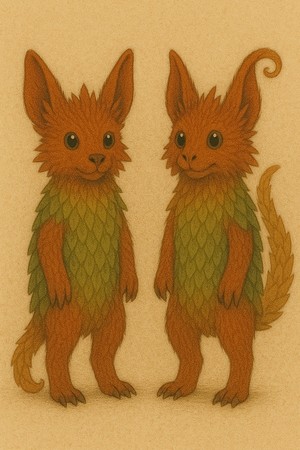} \\[2pt]
            \multicolumn{2}{c}{\textit{Caption this image.}} \\
            Look at the \texttt{[wug]}. & Look at the \texttt{[wugs]}.
        \end{tabular}
        \caption{Vision: training}
        \label{fig:vis-train}
    \end{subfigure}\hfill
    % ---------------- Vision: development ----------------
    \begin{subfigure}[t]{0.48\textwidth}
        \centering
        \begin{tabular}{@{}cc@{}}
            \includegraphics[height=0.95in]{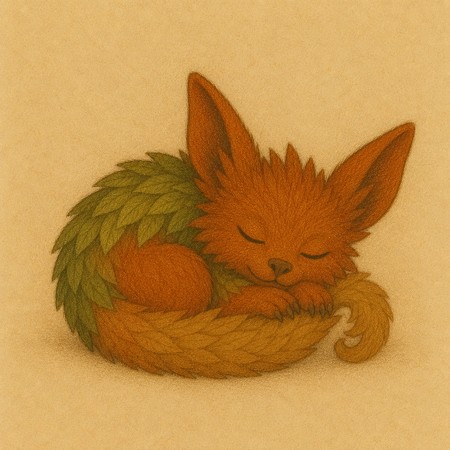} &
            \includegraphics[height=0.95in]{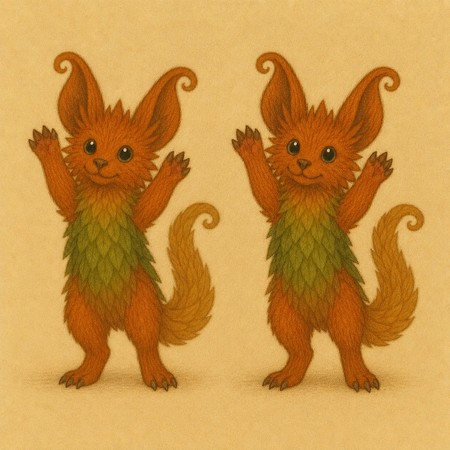} \\[2pt]
            \multicolumn{2}{c}{\textit{What do you see in this picture?}} \\
            I see the \underline{\hspace{1.2em}} $\rightarrow$ \texttt{[wug]} & I see the \underline{\hspace{1.2em}} $\rightarrow$ \texttt{[wugs]}
        \end{tabular}
        \caption{Vision: development}
        \label{fig:vis-dev}
    \end{subfigure}

    \vspace{8pt}

    % ---------------- Language: training ----------------
    \begin{subfigure}[t]{0.48\textwidth}
        \centering
        The \texttt{[wug]} shook its tail before running away.\\
        The \texttt{[wugs]} shook their tails before running away.
        \caption{Language: training}
        \label{fig:lang-train}
    \end{subfigure}\hfill
    % ---------------- Language: development ----------------
    \begin{subfigure}[t]{0.48\textwidth}
        \centering
        The tiny \texttt{[wug]} \{jumps / *jump\}.\\
        The tiny \texttt{[wugs]} \{jump / *jumps\}.
        \caption{Language: development}
        \label{fig:lang-dev}
    \end{subfigure}

    \caption{Examples of training and development stimuli for \texttt{[wug]} and \texttt{[wugs]} in the vision (a--b) and language (c--d) cue conditions. In the vision condition, captions are identical for \texttt{[wug]} and \texttt{[wugs]}, and development items pair held-out images with captions that end right before the novel noun. * denotes the ungrammatical option.}
    \label{fig:stimuli}
\end{figure*}

%\subsection{Methods}
%\label{sec:nwl}

% \km{\@Zach please briefly describe method here: initialization of novel word embeddings, training process (optimizer, lr search---just mention that you did this, with details in appendix, number of samples, dev set)}

\subsection{Training Novel Embeddings}
\label{sec:embed_train}

\paragraph{Initialization} We begin by adding new embeddings, $e_1, e_2$ as new entries in models' embedding ($W_e$) and unembedding ($W_u$) matrices, as well as its tokenizer. We initialize the embeddings with gaussian noise with the mean and standard deviation of 28 singular nouns and their 28 plural counterparts.\footnote{Because these are balanced around equal numbers of singular and plural nouns, the embeddings themselves do not carry any bias towards a particular number} Furthermore, while we use \wug{} and \wugs{} for convenience, the model does not see their orthography, and \textbf{does not} break them into \textit{wug} and \textit{wug+s}. We then freeze the entire model except for these newly added embeddings.
% Our approach to training novel word embeddings is as follows: given a pretrained vision-language model $\mathcal{M}$ with a vocabulary of learned embeddings $E = \{e_0 \ldots e_n\}$, we append a set of novel embeddings $\{x_0 \ldots x_i\}$ to the model vocabulary (in both the embedding matrix $W_E$ and the unembedding matrix $W_U$), where $\{x_0 \ldots x_i\}$ are randomly initialized with gaussian noise around the same subspace of pretrained embeddings:\footnote{In our case, we initialize our novel noun embeddings around a combined mean embedding of 25 singular and 25 plural natural nouns, see Appendix~\ref{}}
% \begin{equation}
% x_j \sim \mathcal{N}(\bar{\mu}, \sigma^2 I), \quad \bar{\mu} = \frac{1}{|S|}\sum_{e \in S} e,
% \end{equation}
% and $S \subset E$ is some subset of natural noun embeddings. We then freeze all model parameters in $\mathcal{M}$ \textit{except} the novel embeddings $\{x_0 \ldots x_i\}$ in $W_E$ and $W_U$, and ensure input and output embeddings are tied so that gradient updates will flow from $W_U$ to $W_E$.

\paragraph{Training} 
As we will see below, the inputs to the model are either a simple sentence without any image, or an image along with an associated text caption. Given an input, we perform training via backpropagation with the cross-entropy loss on the textual part of the stimuli, by only updating $e_1$ and $e_2$---i.e., no other parameter in the model is updated. In our experiments, we primarily report results over 50 seeds (i.e., 50 different initializations) unless stated otherwise. We train for a maximum of 20 epochs, using early stopping on an evaluation set (described briefly below) with a patience of 5. We conducted large scale hyperparameter tuning for learning rate, with details in \Cref{app:hyperparam-sweep} and \Cref{app:imp-resources}.
% After initializing the set of novel embeddings $\{x_0 \ldots x_i\}$, we define a set of training stimuli containing the embeddings, and an additional evaluation set used to assess embeddings learning at each epoch. We then train the unfrozen model parameters in $W_E$ and $W_U$ with BLANK loss for $n$ epochs, halting when evaluation performance does not improve for 5 consecutive epochs.

\paragraph{Cue Conditions} We primarily compare generalization across two types of cues to the novel nouns' number: 1) \textbf{Vision}, where we provide the model with images that depict one or more novel-looking creatures,\footnote{We show in \Cref{sec:img-choice} that our methods are robust to the choice of images, by experimenting with alternatives, like ``blobs on a white background'' or a different imaginary creature.} with a single creature mapped to \wug{} and more than one of the same type mapped to \wugs{}. We create these images by using the OpenAI API,
with prompts shown in \Cref{lst:snarple-prompts,lst:snarple-prompts-pl},
and manually verified them, to maximize the chance that the VLMs have not already seen these images. We pair each image with an associated piece of text (e.g., \textit{Do you see the \wug{}/\wugs{}?}), and use the exact same text caption templates, only differing in the slot occupied by \wug{} or \wugs{}. This way, the only disambiguating signal for number is from the visual modality. We compare this to 2) \textbf{Language}, where the surface form of the text directly disambiguates the nouns' number feature. We do so by pairing the novel nouns with verbs/determiners/quantifiers of the appropriate number (e.g., \textit{The \wug{} \textbf{runs}} vs. \textit{The \wugs{} \textbf{run}}).
For both cue conditions, we use 15 pairs of stimuli for each noun, amounting to a total of 30 stimuli used for training.

\paragraph{Halting Condition} We select the final state of the embeddings in each cue condition differently. In both cases, evaluation is carried out via minimal-pair testing where we compute the percentage of time the model finds acceptable sentence more likely than unacceptable ones. For \textbf{Vision}, we curate a set of held-out images (10 each for singular and plural) and pair them with 6 possible captions (disjoint from the train set), which do not give away the number in the language modality. The captions differ only in terms of \wug{} or \wugs{} (see \Cref{fig:vis-dev}), and the task of the model is to choose the right word given the image (e.g., when the image contains more than one creature we expect the model to prefer \textit{I see the \wugs{}} over \textit{I see the \wug{}}). For \textbf{Language}, we manually curate a development set of 280 minimal pair sentences that test for number agreement (disjoint from the test set). More evaluation sentences can be found in Appendix \ref{app:dev-set-construct}.

\subsection{Evaluation Stimuli}
\label{sec:evalstim}

Following both targeted syntactic evaluation of LMs \citep{marvin-linzen-2018-targeted, gulordava-etal-2018-colorless}, as well as psycholinguistic precedence \citep{bock1992regulating, franck2002subject, arehalli2020neural}, we use minimal pair \textit{agreement attraction} stimuli to quantify number-agreement performance in our experiments. That is, we use declarative sentences with intervening nouns that carry the opposite number feature than that of the subject of the sentence, and therefore ``attract'' the prediction of a verb that agrees with them instead of agreeing with the subject. In the context of both humans and LMs, we generally see a greater number of errors as the number of attractors increases \citep[e.g.,][]{franck2002subject, arehalli2020neural}.
We use the following template, with $n$ = 0,1,2,3 attractors:
\begin{quote}
    \texttt{The [adj] [noun]}$_{\texttt{subj}}$\texttt{ \{[prep] the [noun]$_{\texttt{attr}}$\}}$^n$ \texttt{ [verb]}
\end{quote}
where \texttt{[adj], [noun]$_{\texttt{subj}}$, [prep], [noun]$_{\texttt{attr}}$,} and \texttt{[verb]} denote adjective, subject-noun, preposition, attractor noun, and target verb respectively. We create sentence pairs by sampling disjoint combinations of items from 25 adjectives, 40 nouns, 88 attractor nouns, and 180 verb pairs (all of which are single token),
% \footnote{These reflect filtered sets of words corresponding to one token} 
where each sentence pair has the same noun phrase prefix, and only minimally differs in the verb in terms of its agreement with the subject. We sample 700 sentence pairs per attractor, giving us 2,800 total pairs.
We generate these pairs with real noun subjects, and then replace them with \wug{} and \wugs{} when we evaluate models on number agreement for novel nouns.

\subsection{Models} Our case study focused on the 2B and the 4B version of Qwen3-VL \citep{bai2025qwen3}, both of which are post-trained VLMs. The embedding matrices are tied for both models---i.e., their unembedding and embedding matrices share the same weights. Future work can extend this method to models with untied embeddings.

\section{Behavioral Evidence}
\label{sec:behavioral}

% - Attractor plot with 50 seeds (TODO KM)
% - Dev set: % https://github.com/zstud04/wug-test-interp/blob/main/data/embeddings/eval/training_eval.csv
% - gen constructions: % https://github.com/zstud04/wug-test-interp/blob/main/results/eval/generalization_constructions/Qwen3-VL-4b-Instruct/vision/generalization_constructions_scored.csv

% - Given image does model associate it with the right token?
Our main experiment involves testing the extent to which VLMs are able to generalize the knowledge of the novel nouns' numbers from their exposures to cues in the two different conditions as described in Section \ref{sec:embed_train}. This experiment forms the basis of all subsequent analyses (\Cref{sec:dynamics} and \Cref{sec:mechanistic}). We specifically test models' knowledge of number agreement for our novel nouns by performing minimal pair analyses on agreement-attraction stimuli as mentioned in \cref{sec:evalstim}. We additionally compare these results to those where the subjects of the sentences are real nouns. Since the models have ostensibly encountered these real nouns far more frequently during their training, we naturally expect there to be a gap in the models' performance. For each cue condition, we evaluate models on novel noun embeddings obtained from 50 different training seeds.
% That is, given that the novel words' embeddings have been updated, we then test if models can accurately assign greater probability that agrees with these novel words' number features in sentences where they are subjects. Since these tests involve sentences where the target verb

\begin{figure}[!t]
    \centering
    \includegraphics[width=0.9\columnwidth]{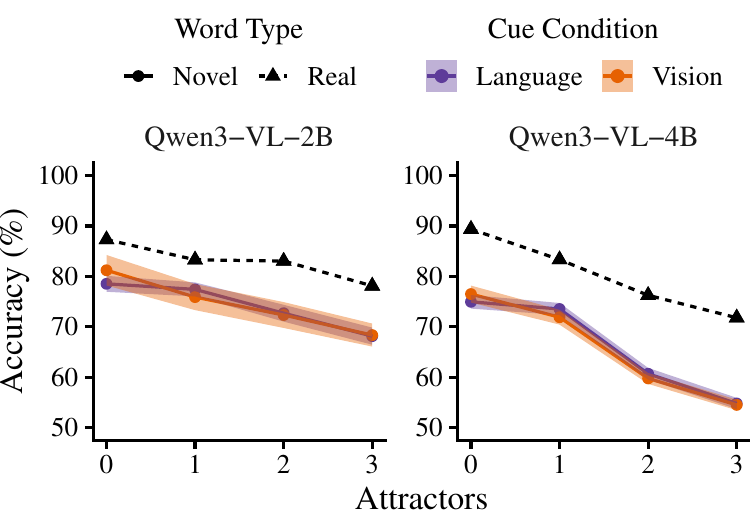}
    \caption{Behavioral accuracy (and 95\% CI) of models on number-agreement stimuli with novel nouns as subjects, across different cue conditions (Language and Vision), and across different intervening attractors (0--3). Black dashed line with triangular points indicates performance on the same stimuli but with real nouns.}
    \label{fig:behavioral-results}
    \vspace{-1em}
\end{figure}

\Cref{fig:behavioral-results} shows these results on both models across different attractors (0--3), cue-conditions (language vs. vision), and subject word types (real vs. novel). As expected, we see a noticeable gap between number agreement performance on real, natural nouns relative to that on novel nouns, and that this performance degrades with increasing number of attractors. At the same time, regardless of the cue condition, we also see that models are above chance at number agreement for all attractors, suggesting successful generalization. This is especially striking in the vision condition---our results show that even when number cannot be diagnosed by textual co-occurrence (as in the `Language' case), VLMs are able to demonstrate behavior compatible with successful generalization, and by extension, successful encoding of \textit{number}. 
% Interestingly, the 4B model---although it is unexpectedly worse than its 2B counterpart---shows similar behavior for embeddings learned from both modalities, while the 2B model favors the language cue condition slightly over its vision counterpart. \km{unsure if worth pointing out}

% That is, given that a model has encountered the nouns \wug{} and \wugs{} in the language (e.g., \textit{The \wug{}/\wugs{} is/are over there.}) or vision cue conditions (e.g., [image of one/two creatures] + \textit{Do you see the \wug{}/\wugs{}?}), to what extent can the model make accurate judgments on number agreement stimuli?

% \km{We test models after performing novel word learning experiments on text-stimuli that tests for number agreement, with intervening nouns with the opposite number to the subject. We compare natural vs. novel and report results in fig 2. We see that novel is noticeably worse than natural but all are above chance (50\%). Language and vision are especially similar in 4B. overall, models are capable of generalizing even when cues come from vision.}

% results on attractors for 2B and 4B for both L and V. Both learned well, and drop as expected with more attractors.

% \km{Is there anything here we can add to speculate if it's at all possible that this is shown without any abstraction? I suppose the mere fact that this happens suggests abstraction-compatible behavior, though}

\section{Representational Dynamics of Learning}
\label{sec:dynamics}

Having shown that our VLMs are able to generalize number agreement for novel words learned from both vision and language cues, we now move onto analyses that characterize the internal dynamics of these words' representations during learning. 
Is there a systematic pattern to how these representations change?
To this end, we follow \citet{misra-and-kim-2023-catabs}, and track the movement of the novel word embeddings in low dimensional subspaces that capture abstractions relevant to number agreement. 
We report these results for the 2B model here, and include results for the 4B model in \Cref{app:2b-results}.

\subsection{Methods}
We start by performing a 2-dimensional Principal Component Analysis (PCA) on a subset of the embedding layer of our models consisting of: 1) pairs of real-world nouns whose number information is well-known (e.g., \textit{dog--dogs}, \textit{car-cars}, etc.); and 2) the initial and final states of the novel word pairs from all 50 seeds used in previous experiment. We take the randomly initialized versions of these novel words to be the initial state and their states at the end of each training run to be their respective final states. We repeat this for embeddings learned from language as well as vision cues. To obtain our set of real nouns, we query WordNet \citep{miller1995wordnet} and sample 500 singular nouns and their plural counterparts (amounting to a total of 1000 nouns), which exist in the single-token vocabulary of our models' tokenizers.\footnote{It is important to note that the embeddings for these real words have not changed, since we only backpropagate over the novel words' embeddings.} Upon reducing our embeddings to 2 dimensions, we then connect each novel word's initial state to its final state using an arrow, and then visualize this movement with respect to the representations of the real nouns. 

To quantify this movement, we compute the scalar projection of each of our novel word embeddings (initial and final states separately) onto the direction that captures noun-number. To obtain the number direction, we took the difference between the vector formed by averaging the embeddings of all our singular nouns ($v_{\textsf{sg}}$) and that formed from the average embedding of all our plural nouns ($v_{\textsf{pl}}$). For a given embedding of a novel noun ($e_{novel}$), we then compute the scalar projection, and subsequently the movement as:
\begin{align}
    % d = v_{\textsf{sg}} - v_{\textsf{pl}}
    \textrm{proj} &= e_{\textsf{novel}} . \frac{v_{\textsf{sg}} - v_{\textsf{pl}}}{||v_{\textsf{sg}} - v_{\textsf{pl}}||}\\
    \textrm{movement} &= \textrm{proj}_{\textsf{final}} - \textrm{proj}_{\textsf{initial}}
\end{align}
That is, we quantify movement as the change in the projection of a novel noun embedding (final - initial) onto the singular-plural direction. Therefore, novel nouns that are supposed to be treated as singular should show positive movement, and those that are supposed to be treated as plural should show negative movement. This method has been used in the past to measure bias \citep{bolukbasi2016man} as well as encoding of semantic features in vector space models \citep{grand2022semantic}.
% is similar to \citet{grand2022semantic}'s semantic projection as well as \citet{bolukbasi2016man}'s 

\subsection{Results} 
\begin{figure}[!t]
    \centering
    \includegraphics[width=0.8\linewidth]{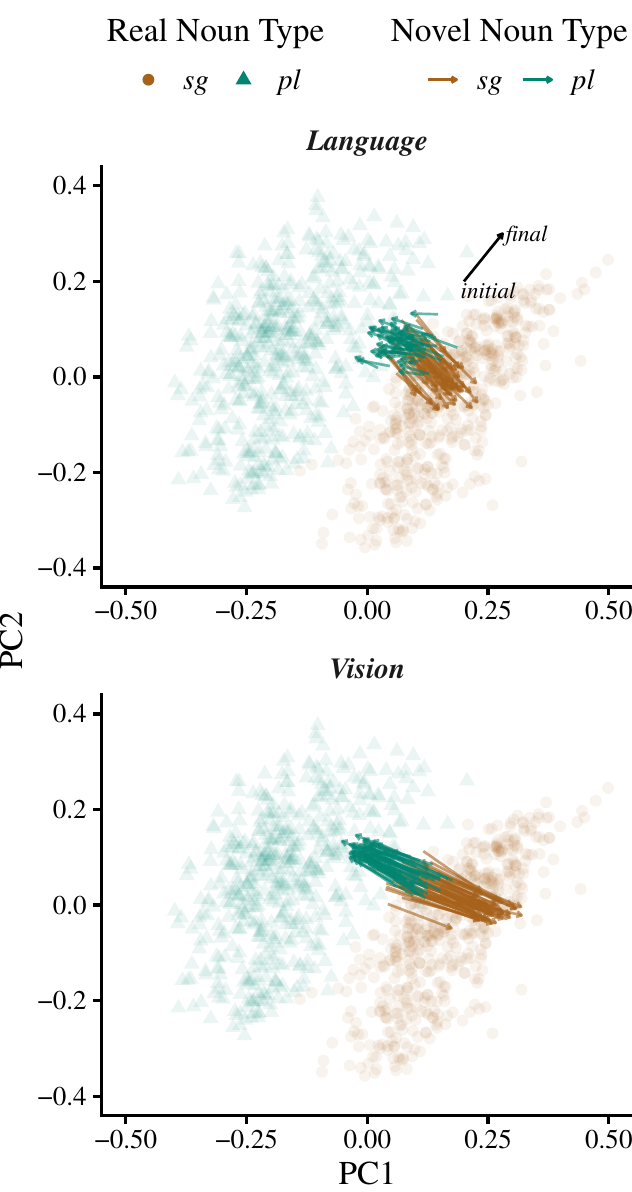}
    \caption{Movement (shown using arrows) of the embedding states of novel words when analyzed using a 2D PCA fit on embeddings of real singular (\textit{sg}) and plural (\textit{pl}) nouns (e.g., \textit{dogs, chairs, blocks,} etc., $N$=500 each) in the Qwen3-VL-2B model across both cue conditions (Language and Vision). Colors indicate grammatical number for real and novel nouns.}
    \label{fig:movement-pca-2b}
    % \vspace{-1em}
\end{figure}

\Cref{fig:movement-pca-2b} visualizes the first two principal components of the model embeddings across both cue conditions---language and vision. We first see that real nouns are organized according to their number features---i.e., the set of 500 singular nouns end up clumping together, and similarly so does the set of their plural counterparts. Then, regardless of the cue condition, we see non-trivial movement of the novel nouns towards their respective directions. That is, novel singular and plural nouns across both modalities move towards the regions occupied by real nouns with the respective number features. 
% (i.e., singular to singular, and plural to plural).

% \paragraph{Method} 
% Unambiguous nouns sampled from wordnet, in model's single token vocabulary (both sg and pl). PCA into 2 dimensions along with start and end states of the novel word (50 seeds). Plot for 4B (main). For movement, get centroid of known sg and pl regions, then do vector projection. Movement = how much did it move to the right direction. All 2B plots in appendices.

% \begin{table}[!t]
% \centering
% \resizebox{0.8\columnwidth}{!}{
% \begin{tabular}[t]{lcc}
% \toprule
% \textbf{Cue Condition} & \textit{\textbf{sg}} & \textit{\textbf{pl}}\\
% \midrule
% Language & 0.092\textsubscript{$\pm$ 0.006} & -0.061\textsubscript{$\pm$ 0.006}\\
% Vision & 0.067\textsubscript{$\pm$ 0.007} & -0.040\textsubscript{$\pm$ 0.004}\\
% \bottomrule
% \end{tabular}
% }
% \caption{Mean movement by Cue Condition and number. Movement is calculated as the difference in the projection of the novel word embeddings' final and initial states onto the singular-plural direction, computed as the vector difference of real singular and plural nouns. Movements are significantly different than 0 ($p<$.001).}
% \label{tab:mean-movement}
% \vspace{-1em}
% \end{table}

\begin{table}[!t]
    \centering
    \resizebox{0.8\columnwidth}{!}{
    \begin{tabular}{lcc}
        \toprule
        \textbf{Cue Condition} & \textbf{\textit{sg}} & \textbf{\textit{pl}} \\
        \midrule
        Language & 0.088\textsubscript{$\pm$ 0.005} & -0.051\textsubscript{$\pm$ 0.004} \\
        Vision   & 0.164\textsubscript{$\pm$ 0.011} & -0.118\textsubscript{$\pm$ 0.007} \\
        \bottomrule
    \end{tabular}
    }
    \caption{Mean movement by Cue Condition and Number, for Qwen3-VL-2B. Movement is calculated as the difference in the projection of the novel word embeddings' final and initial states onto the singular-plural direction, computed as the vector difference of real singular and plural nouns. Movements are significantly different than 0 ($p<$.001).}
    \label{tab:mean-movement}
    \vspace{-1em}
\end{table}

\Cref{tab:mean-movement} shows the average movement (averaged across 50 seeds) of the novel embeddings for both cue-conditions as well as for both types of number features. First, all movements are in their intended directions---we see positive movements for novel embeddings that are supposed to be singular and negative movements for embeddings that are supposed to be plural ($p <$.001 for both). Overall, even though there is no diagnostic information in the language component of the visual cue, we still see non-trivial movement during learning towards the space of desirable exemplars---even more so than in the language condition. This suggests that representational movement towards regions inhabited by exemplars that bear the target feature (here, \textit{number}) underlies abstraction compatible behavior in VLMs, regardless of modality.

% \km{If there's no space we can put the movement metric in appendix and only show PCA}

\section{Mechanistic Evidence}
\label{sec:mechanistic}

Results from the previous experiment shed light on what happens inside the embeddings of novel words as they are being integrated within the models' existing embedding layer. How is this information transmitted from the input embedding to the rest of the model in a manner that enables it to produce the right output? For this, we turn to a slew of modern mechanistic interpretability methods. To us, the goal of these methods is to describe the internal mechanisms that underlie an abstraction. If these mechanisms are discovered \textit{before} our novel nouns (e.g., \wug{} and \wugs{}) are added to the model, then to what extent do they make consistent predictions about novel words \textit{after} they are acquired? This also lets us shed light on the generalization capabilities of these methods in a manner that goes beyond stimuli that the models have seen at the time of mechanism discovery.

\subsection{Methods}

We rely on methods that involve counterfactual interventions on the models' internal components in a manner that manipulates the model output \citep{arora-etal-2024-causalgym, boguraev-etal-2025-causal, mueller-etal-2026-quest}. Our methods follow the framework of \citet{arora-etal-2024-causalgym}, where each method involves pairs of stimuli, \textit{base} and \textit{source}, with their corresponding next word labels, $y_b$ and $y_s$. We use these stimuli to intervene on a model component, $f$, by taking its value from the \textit{source} and applying it to \textit{base} using some sort of a method-dependent transformation to produce $f^*$. Insofar as this intervention succeeds, we should observe a change in the intervened model's output probabilities $p_{f \leftarrow f^*}$ that is compatible with the output when the input was the \textit{source} stimulus, relative to the original model probabilities $p_f$.
We measure the efficacy of the intervention using the log odds-ratio which compares the probabilities of the source and base next-word labels before and after intervention:
\begin{align*}
     \textsf{Odds}&(f, f^*, \langle b, s, y_b, y_s \rangle)\nonumber\\
     &= \log \left(\frac{p_f(y_b \mid b)}{p_f(y_s \mid b)}\cdot\frac{p_{f \leftarrow f^*}(y_s \mid b)}{p_{f \leftarrow f^*}(y_b \mid b)} \right)
\end{align*}
We validate our methods by fitting/finding interventions on a train set (except for one of the methods, which is unsupervised) and evaluating on a held-out test set, and in this case, we only use stimuli with real words, since we perform these interventions before novel words are added to the model. Our train and test stimuli are sampled using our attractor stimuli method described in \Cref{sec:evalstim}. In particular, we split these stimuli (keeping singular and plural ones balanced) such that the set of subjects, attractor nouns, and target verb pairs are completely disjoint between train ($N$=700 per attractor) and test ($N$=700 per attractor). We evaluate using the average log odds-ratio computed over our test set. 

Given the aforementioned analysis design, we run experiments on four intervention methods:\footnote{Detailed description of each method is provided in \Cref{app:intervention-details}.} 

\begin{figure*}[!ht]
    \centering
    \includegraphics[width=0.8\linewidth]{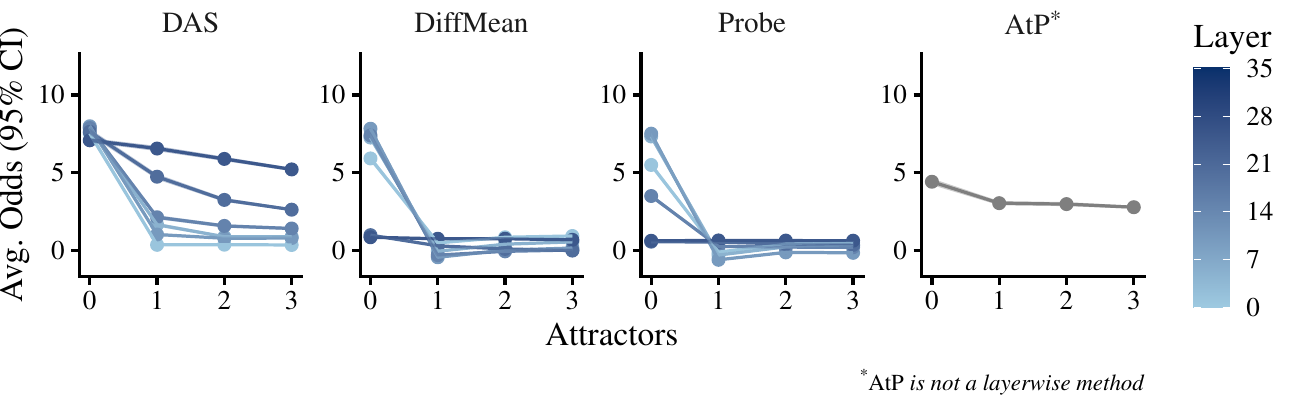}
    \caption{Avg. Odds across model layers (when applicable) for the Qwen3-VL-2B model on agreement stimuli with real nouns as subjects. Higher values mean greater causal effect.}
    \label{fig:real-interp-2b}
\end{figure*}

\begin{figure*}[!ht]
    \centering
    \includegraphics[width=0.8\linewidth]{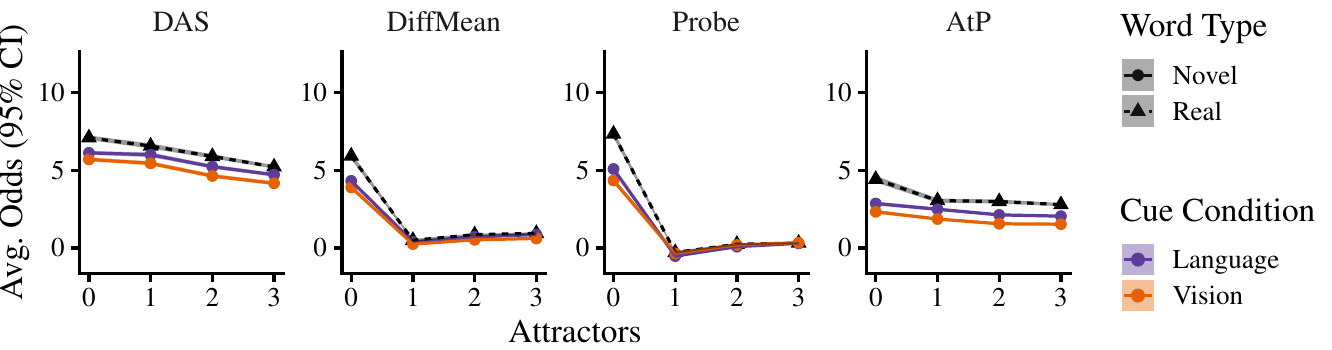}
    \caption{Avg. Odds for the Qwen3-VL-2B model on agreement stimuli with subjects that are real vs. novel nouns acquired from both types of Cue conditions. Results shown for layer with best overall avg. odds chosen on results on stimuli with real nouns as subjects. We see generally high agreement in the results across both cue-conditions.}
    \label{fig:novel-vs-real-interp-2b}
    \vspace{-1em}
\end{figure*}

% 1) \textbf{Distributed Alignment Search} \citep[DAS][]{geiger2021causal, geiger2024finding}, which performs interventions by learning rotations that maximizes the likelihood of the counterfactual predictions on performing the intervention; 2) \textbf{}

% Specifically, we use the following methods: 

\paragraph{Distributed Alignment Search (DAS)}   This method \citep{geiger2021causal, geiger2024finding} learns a subspace in a model's activations in a manner that maximizes the likelihood of a given counterfactual completion token on performing interventions. We restrict this rotation to be 1-dimensional, and apply it layer-wise. This method is by definition sensitive to changes in the full model behavior.

\paragraph{DiffMean} This method \citep{marks2023geometry} intervenes by adding an offset vector (or its negative) with a coefficient ($\pm\alpha \mathbf{v}$) that is computed using the difference in the average vector of activations (at each layer) per class (here, singular and plural). Since this method is unsupervised, we directly applied it to the test set, and use $\alpha=50$. 

\paragraph{Probe} We fit linear probes \citep{alain2016understanding, ettinger2016probing}---in our case, using logistic regression---on model activations (across layers) for the train set and then perform interventions by adding the learned weights along with a coefficient  ($\pm\alpha \mathbf{w}$, where $\mathbf{w}$ is the learned weight vector of logistic regression, and $\alpha$ is set to 50).
% This method does not optimize for causality.

\paragraph{Attribution Patching (AtP)} This method \citep{nanda2023attribution} approximates the individual effects of a set of model components using the gradients computed from a single backwards pass over those components.  In our case, we compute these gradients with respect to the logit difference of the correct versus incorrect sentence completion (in short, estimating the components with greatest contribution in producing the model completion "\textit{The wugs \textbf{are}}" versus "\textit{The wugs \textbf{is}}"). To produce a causal effect in the model, we then extract the top-$k$ activations (as ranked by the attribution method) in the forward pass for a given singular prompt and patch those activations into the forward pass for its plural complement (and vice versa).

When applicable, we apply the above methods layer-wise\footnote{We apply them at every 5th layer and additionally the last layer for each model} and on the last token position prior to the model prediction.
% \footnote{We also tried on other token positions but found greatest causal effect on the penultimate position.}
To evaluate generalization to novel words, we select the layer (when applicable) with the best average log odds-ratio. Then, after adding our novel nouns into the embedding layer, we further evaluate the method on our test set now with the subject nouns appropriately replaced with our novel nouns. We repeat this process for both language and vision cue conditions.

\subsection{Results}

\Cref{fig:real-interp-2b} shows results from our validation experiments, where we report Avg. Odds across methods, and when applicable, layers, on stimuli with real nouns as subjects. We see that the Avg. Odds across \textit{all} methods are above 0, indicating qualitatively successful interventions. Similarly to \citet{arora-etal-2024-causalgym}, we find DAS to be most causally efficacious, with comparatively higher Avg. Odds across attractors and layers.

Turning now to results reported in \Cref{fig:novel-vs-real-interp-2b}, we investigate how well these methods incorporate information from novel nouns that have been acquired \textit{after} the mechanisms were discovered, across both cue conditions (Language and Vision). These results only include Avg. Odds from layers that were selected to be the best in the previous analysis on real nouns. We again observe Avg. Odds that are generally above 0, across all methods and for both cue conditions. More interestingly, with the exception of DAS for 0 attractors, we see no visual difference between the Language and Vision cue conditions. To test this further, we fit a linear mixed effects model on all our results (across all layers, and both models), where we predict the avg. odds using cue condition, attractors, model, and interpretability method as fixed effects, and layer as random effects. Here, we find no significant effect of the cue condition ($\beta_{cue}$=-0.068, $p$=.65).\footnote{For full results, see \Cref{sec:stats}.} Overall, this suggests that linguistic evidence is not privileged in mechanisms responsible for number-agreement behavior in models, and that words whose number information is learned exclusively from vision are integrated in the same way as those learned from explicit linguistic evidence.
% Turning now to results on how well these methods incorporate information from novel nouns \textit{after} they have 

% Multiple different methods -- DAS, Logistic Regression, Circuits...?

% 

\section{General Discussion and Conclusion}
\label{sec:discussion}

% \paragraph{VLMs learn powerful cross-modal abstractions...}
% They are able to learn about number from both linguistic and visual cues, especially even if all they know about the novel word is from visual evidence \textit{alone}.

% \paragraph{...which guide further learning of new exemplars...} Learning is characterized by regions of known singular and plural nouns which serve as basins that `attract' novel exemplars, regardless of modality. For instance, novel words that are exposed with cues that denote sg move closer to regions occupied by other singular nouns, and similarly for novel words exposed with pl-cues. This is observed even when the cues that disambiguate the number feature are non syntactic and in fact come from a different modality (e.g., vision!)

% \paragraph{...and are accounted for by a common causal mechanism} 
% we find that this behavior of the model is also accounted for by the same mechanism that explains number-agreement in these models, even though the mechanism itself was discovered \textit{before} the novel words were learned (i.e., before their embedding is updated). behavior is better accounted by methods grounded in causal abstractions .
Using analyses of model behavior, internal representational dynamics, and causal mechanisms, we find that VLMs can demonstrate cross-modal generalization of number agreement. That is, they can learn about a novel noun's number from both linguistic and visual cues, especially even if all they know about the noun is from visual evidence \textit{alone}. This learning is characterized by emergent abstraction of the number feature, which is learned for words already present in the model embeddings and which guide the learning of this information for these new nouns. Finally, the internal mechanisms responsible for number agreement in the model are able to account for data the model acquires \textit{after} they have been discovered, and they treat nouns learned from language vs. vision alone in the same way.

These findings indicate that emergent abstractions in data-driven learners like VLMs are more flexible than their initial mode of acquisition---LMs primarily learn about number from linguistic cues: the pairing of nouns with commonly co-occurring verbs. When they eventually learn to map words to images (and vice-versa) during VLM training, their distributional evidence is further bolstered, but it is bolstered for nouns for which they already have strong evidence (for number). Our results show that even when they learn nouns \textit{only} from vision, the tested VLMs' learned linguistic abstractions are \textit{prepared} to accommodate this new knowledge in a manner that is seemingly indistinguishable from nouns that are acquired only from text. 

Our findings also speak to a challenge raised by several commentators on radical exemplar accounts of acquisition \citep{ambridge2020abstractions, lieven2020similarity, demuth2020exemplar}. \citet{lieven2020similarity} argued that for such accounts to move forward, they must clearly specify the mechanisms by which analogy and similarity are computed (among other points). \citet{demuth2020exemplar} further argue that exemplar-based learning requires both a procedure for segmenting experience into units that can be compared, and a similarity function for comparing them---for a learner to acquire a language, both of these likely require abstraction. Our analyses offer \textit{one} case study of what such a similarity function might look like. In our analyses, the similarity that guides learning is directly observable in the (V)LMs' representational space. Novel nouns move towards regions of embedding space inhabited by embeddings of real nouns that share their number feature, and this movement can be quantified as a projection onto a direction defined by those real noun embeddings, which are themselves abstractions aggregated over exposures to individual items (usages of nouns in training). Importantly, this similarity is not computed \textit{merely} over the input's actual surface-form---our novel nouns share no orthography with any real noun, and in the vision condition, nor do they share any linguistic context that specifically diagnoses their number. Instead, this similarity is in a learned representational space that integrates evidence across modalities. Furthermore, the resulting generalization operates at the level of the number category rather than individual exemplars: the same causal subspaces that support agreement for real nouns also support it for novel ones.

Overall, our results suggest that generalization in these models might be learned \textit{from} surface-form co-occurrence, but that they can generalize beyond this knowledge and show evidence of abstraction-compatible behavior. Abstract and item-based learning might not be so different from each other after all \citep{ambridge2020against, ambridge2020abstractions, misra-and-kim-2023-catabs}.

\section*{Limitations}

% Focused + only on English -- but we do it in-depth in a holistic manner. Future works can adapt our models to explore multiple phenomena.
While we have, in as much detail as possible, provided in-depth evidence for cross-modal generalization for number-agreement, there are several ways in which our study is limited. In particular, our focus is only on two VLMs from the same family with largely lower parameter sizes, with limited number of images, and on a single language (English). Overall, while we cannot make strong claims about VLMs as a class, our results, along with several others who have conducted studies on a single model \citep[][etc.]{petty2022position, misra-and-kim-2023-catabs, jian-manning-2026-humans} suggest that it is, in principle, possible for abstractions to emerge via item-based learning.

Furthermore, our experiments only use a single method to conduct novel word learning---this was a deliberate choice, because many alternatives \citep{teehan2024college, wang-etal-2025-rapid} often invoke a separate model or an atypical training process (e.g., meta learning based fine-tuning). Using these would have prevented us from drawing conclusions about the target model (cf. the idiosyncrasies of a particular method).

% \section*{Acknowledgments}

% Bibliography entries for the entire Anthology, followed by custom entries
%\bibliography{anthology,custom}
% Custom bibliography entries only
\bibliography{custom,refs,kanishka}

@inproceedings{bolukbasi2016man,
 author = {Bolukbasi, Tolga and Chang, Kai-Wei and Zou, James and Saligrama, Venkatesh and Kalai, Adam T},
 booktitle = {Advances in Neural Information Processing Systems},
 editor = {D. Lee and M. Sugiyama and U. Luxburg and I. Guyon and R. Garnett},
 pages = {},
 publisher = {Curran Associates, Inc.},
 title = {Man is to Computer Programmer as Woman is to Homemaker? Debiasing Word Embeddings},
 url = {https://proceedings.neurips.cc/paper_files/paper/2016/file/a486cd07e4ac3d270571622f4f316ec5-Paper.pdf},
 volume = {29},
 year = {2016}
}

@inproceedings{boguraev-etal-2025-causal,
    title = "Causal Interventions Reveal Shared Structure Across {E}nglish Filler{--}Gap Constructions",
    author = "Boguraev, Sasha  and
      Potts, Christopher  and
      Mahowald, Kyle",
    editor = "Christodoulopoulos, Christos  and
      Chakraborty, Tanmoy  and
      Rose, Carolyn  and
      Peng, Violet",
    booktitle = "Proceedings of the 2025 Conference on Empirical Methods in Natural Language Processing",
    month = nov,
    year = "2025",
    address = "Suzhou, China",
    publisher = "Association for Computational Linguistics",
    url = "https://aclanthology.org/2025.emnlp-main.1271/",
    doi = "10.18653/v1/2025.emnlp-main.1271",
    pages = "25021--25042",
    ISBN = "979-8-89176-332-6"
}

@inproceedings{hu-etal-2020-systematic,
    title = "A Systematic Assessment of Syntactic Generalization in Neural Language Models",
    author = "Hu, Jennifer  and
      Gauthier, Jon  and
      Qian, Peng  and
      Wilcox, Ethan  and
      Levy, Roger P.",
    editor = "Jurafsky, Dan  and
      Chai, Joyce  and
      Schluter, Natalie  and
      Tetreault, Joel",
    booktitle = "Proceedings of the 58th Annual Meeting of the Association for Computational Linguistics",
    month = jul,
    year = "2020",
    address = "Online",
    publisher = "Association for Computational Linguistics",
    url = "https://aclanthology.org/2020.acl-main.158/",
    doi = "10.18653/v1/2020.acl-main.158",
    pages = "1725--1744"
}

@inproceedings{mueller-etal-2020-cross,
    title = "Cross-Linguistic Syntactic Evaluation of Word Prediction Models",
    author = "Mueller, Aaron  and
      Nicolai, Garrett  and
      Petrou-Zeniou, Panayiota  and
      Talmina, Natalia  and
      Linzen, Tal",
    editor = "Jurafsky, Dan  and
      Chai, Joyce  and
      Schluter, Natalie  and
      Tetreault, Joel",
    booktitle = "Proceedings of the 58th Annual Meeting of the Association for Computational Linguistics",
    month = jul,
    year = "2020",
    address = "Online",
    publisher = "Association for Computational Linguistics",
    url = "https://aclanthology.org/2020.acl-main.490/",
    doi = "10.18653/v1/2020.acl-main.490",
    pages = "5523--5539"
}

@inproceedings{finlayson-etal-2021-causal,
    title = "Causal Analysis of Syntactic Agreement Mechanisms in Neural Language Models",
    author = "Finlayson, Matthew  and
      Mueller, Aaron  and
      Gehrmann, Sebastian  and
      Shieber, Stuart  and
      Linzen, Tal  and
      Belinkov, Yonatan",
    editor = "Zong, Chengqing  and
      Xia, Fei  and
      Li, Wenjie  and
      Navigli, Roberto",
    booktitle = "Proceedings of the 59th Annual Meeting of the Association for Computational Linguistics and the 11th International Joint Conference on Natural Language Processing (Volume 1: Long Papers)",
    month = aug,
    year = "2021",
    address = "Online",
    publisher = "Association for Computational Linguistics",
    url = "https://aclanthology.org/2021.acl-long.144/",
    doi = "10.18653/v1/2021.acl-long.144",
    pages = "1828--1843"
}

@inproceedings{geiger2021causal,
 author = {Geiger, Atticus and Lu, Hanson and Icard, Thomas and Potts, Christopher},
 booktitle = {Advances in Neural Information Processing Systems},
 editor = {M. Ranzato and A. Beygelzimer and Y. Dauphin and P.S. Liang and J. Wortman Vaughan},
 pages = {9574--9586},
 publisher = {Curran Associates, Inc.},
 title = {Causal Abstractions of Neural Networks},
 url = {https://proceedings.neurips.cc/paper_files/paper/2021/file/4f5c422f4d49a5a807eda27434231040-Paper.pdf},
 volume = {34},
 year = {2021}
}

@inproceedings{arora-etal-2024-causalgym,
    title = "{C}ausal{G}ym: Benchmarking causal interpretability methods on linguistic tasks",
    author = "Arora, Aryaman  and
      Jurafsky, Dan  and
      Potts, Christopher",
    editor = "Ku, Lun-Wei  and
      Martins, Andre  and
      Srikumar, Vivek",
    booktitle = "Proceedings of the 62nd Annual Meeting of the Association for Computational Linguistics (Volume 1: Long Papers)",
    year = "2024",
    address = "Bangkok, Thailand",
    publisher = "Association for Computational Linguistics",
    url = "https://aclanthology.org/2024.acl-long.785/",
    doi = "10.18653/v1/2024.acl-long.785",
    pages = "14638--14663"
}

@inproceedings{marks2025sparse,
  title={Sparse feature circuits: Discovering and editing interpretable causal graphs in language models},
  author={Marks, Samuel and Rager, Can and Michaud, Eric and Belinkov, Yonatan and Bau, David and Mueller, Aaron},
  booktitle={International Conference on Learning Representations},
  volume={2025},
  pages={23888--23923},
  year={2025},
  url={https://proceedings.iclr.cc/paper_files/paper/2025/hash/3ba4d47a83e498c2b1a0868cba20f6de-Abstract-Conference.html}
}

@article{hewitt2025we,
  title={We can't understand {AI} using our existing vocabulary},
  author={Hewitt, John and Geirhos, Robert and Kim, Been},
  journal={arXiv preprint arXiv:2502.07586},
  year={2025}
}

@inproceedings{
teehan2024college,
title={Co{LLEG}e: Concept Embedding Generation for Large Language Models},
author={Ryan Teehan and Brenden Lake and Mengye Ren},
booktitle={First Conference on Language Modeling},
year={2024},
url={https://openreview.net/forum?id=Fkr1yVUb9G}
}

@article{misra2024generating,
  title={Generating novel experimental hypotheses from language models: A case study on cross-dative generalization},
  author={Misra, Kanishka and Kim, Najoung},
  journal={arXiv preprint arXiv:2408.05086},
  year={2026}
}

@inproceedings{wang-etal-2025-rapid,
    title = "Rapid Word Learning Through Meta In-Context Learning",
    author = "Wang, Wentao  and
      Jiang, Guangyuan  and
      Linzen, Tal  and
      Lake, Brenden",
    editor = "Christodoulopoulos, Christos  and
      Chakraborty, Tanmoy  and
      Rose, Carolyn  and
      Peng, Violet",
    booktitle = "Proceedings of the 2025 Conference on Empirical Methods in Natural Language Processing",
    month = nov,
    year = "2025",
    address = "Suzhou, China",
    publisher = "Association for Computational Linguistics",
    url = "https://aclanthology.org/2025.emnlp-main.1631/",
    doi = "10.18653/v1/2025.emnlp-main.1631",
    pages = "32038--32073",
    ISBN = "979-8-89176-332-6"
}

@article{ambridge2020abstractions,
  title={{Abstractions made of exemplars or ‘You’re all right, and I’ve changed my mind’: Response to commentators}},
  author={Ambridge, Ben},
  journal={First Language},
  volume={40},
  number={5-6},
  pages={640--659},
  year={2020},
  publisher={Sage Publications Sage UK: London, England}
}

@inproceedings{petty2022position,
  author    = {Petty, Jackson and Wilson, Michael and Frank, Robert},
  title     = {Do Language Models Learn Position-Role Mappings?},
  booktitle = {Proceedings of the 46th Annual Boston University Conference
               on Language Development},
  editor    = {Gong, Ying and Kpogo, Felix},
  year      = {2022},
  pages     = {657--671},
  address   = {Somerville, MA},
  publisher = {Cascadilla Press}
}

@article{mccoy2024embers,
  title={Embers of autoregression show how large language models are shaped by the problem they are trained to solve},
  author={McCoy, R Thomas and Yao, Shunyu and Friedman, Dan and Hardy, Mathew D and Griffiths, Thomas L},
  journal={Proceedings of the National Academy of Sciences},
  volume={121},
  number={41},
  pages={e2322420121},
  year={2024},
  publisher={National Academy of Sciences}
}

@article{ambridge2015ubiquity,
  title={{The ubiquity of frequency effects in first language acquisition}},
  author={Ambridge, Ben and Kidd, Evan and Rowland, Caroline F and Theakston, Anna L},
  journal={Journal of child language},
  volume={42},
  number={2},
  pages={239--273},
  year={2015},
  publisher={Cambridge University Press}
}

@article{ambridge2020against,
  title={{Against stored abstractions: A radical exemplar model of language acquisition}},
  author={Ambridge, Ben},
  journal={First Language},
  volume={40},
  number={5-6},
  pages={509--559},
  year={2020},
  publisher={Sage Publications Sage UK: London, England}
}

@article{wilson-etal-2023-abstract,
    title = "How Abstract Is Linguistic Generalization in Large Language Models? Experiments with Argument Structure",
    author = "Wilson, Michael  and
      Petty, Jackson  and
      Frank, Robert",
    journal = "Transactions of the Association for Computational Linguistics",
    volume = "11",
    year = "2023",
    address = "Cambridge, MA",
    publisher = "MIT Press",
    url = "https://aclanthology.org/2023.tacl-1.78/",
    doi = "10.1162/tacl_a_00608",
    pages = "1377--1395"
}

@inproceedings{kim-smolensky-2021-testing,
    title = "Testing for Grammatical Category Abstraction in Neural Language Models",
    author = "Kim, Najoung  and
      Smolensky, Paul",
    editor = "Ettinger, Allyson  and
      Pavlick, Ellie  and
      Prickett, Brandon",
    booktitle = "Proceedings of the Society for Computation in Linguistics 2021",
    month = feb,
    year = "2021",
    address = "Online",
    publisher = "Association for Computational Linguistics",
    url = "https://aclanthology.org/2021.scil-1.59/",
    pages = "467--470"
}

@article{lampinen2017one,
  title={One-shot and few-shot learning of word embeddings},
  author={Lampinen, Andrew K and McClelland, James L},
  journal={arXiv preprint arXiv:1710.10280},
  year={2017},
  url={https://arxiv.org/abs/1710.10280}
}

@inproceedings{lasri-etal-2022-bert,
    title = "Does {BERT} really agree? Fine-grained Analysis of Lexical Dependence on a Syntactic Task",
    author = "Lasri, Karim  and
      Lenci, Alessandro  and
      Poibeau, Thierry",
    editor = "Muresan, Smaranda  and
      Nakov, Preslav  and
      Villavicencio, Aline",
    booktitle = "Findings of the Association for Computational Linguistics: ACL 2022",
    month = may,
    year = "2022",
    address = "Dublin, Ireland",
    publisher = "Association for Computational Linguistics",
    url = "https://aclanthology.org/2022.findings-acl.181/",
    doi = "10.18653/v1/2022.findings-acl.181",
    pages = "2309--2315"
}

@inproceedings{lasri-etal-2022-probing,
    title = "Probing for the Usage of Grammatical Number",
    author = "Lasri, Karim  and
      Pimentel, Tiago  and
      Lenci, Alessandro  and
      Poibeau, Thierry  and
      Cotterell, Ryan",
    editor = "Muresan, Smaranda  and
      Nakov, Preslav  and
      Villavicencio, Aline",
    booktitle = "Proceedings of the 60th Annual Meeting of the Association for Computational Linguistics (Volume 1: Long Papers)",
    month = may,
    year = "2022",
    address = "Dublin, Ireland",
    publisher = "Association for Computational Linguistics",
    url = "https://aclanthology.org/2022.acl-long.603/",
    doi = "10.18653/v1/2022.acl-long.603",
    pages = "8818--8831"
}

@article{elman1991distributed,
  title={Distributed representations, simple recurrent networks, and grammatical structure},
  author={Elman, Jeffrey L},
  journal={Machine learning},
  volume={7},
  number={2},
  pages={195--225},
  year={1991},
  publisher={Springer},
  url={https://link.springer.com/article/10.1007/bf00114844}
}

@inproceedings{lakretz-etal-2019-emergence,
    title = "The emergence of number and syntax units in {LSTM} language models",
    author = "Lakretz, Yair  and
      Kruszewski, German  and
      Desbordes, Theo  and
      Hupkes, Dieuwke  and
      Dehaene, Stanislas  and
      Baroni, Marco",
    editor = "Burstein, Jill  and
      Doran, Christy  and
      Solorio, Thamar",
    booktitle = "Proceedings of the 2019 Conference of the North {A}merican Chapter of the Association for Computational Linguistics: Human Language Technologies, Volume 1 (Long and Short Papers)",
    month = jun,
    year = "2019",
    address = "Minneapolis, Minnesota",
    publisher = "Association for Computational Linguistics",
    url = "https://aclanthology.org/N19-1002/",
    doi = "10.18653/v1/N19-1002",
    pages = "11--20"
}

@inproceedings{wei-etal-2021-frequency,
    title = "Frequency Effects on Syntactic Rule Learning in Transformers",
    author = "Wei, Jason  and
      Garrette, Dan  and
      Linzen, Tal  and
      Pavlick, Ellie",
    editor = "Moens, Marie-Francine  and
      Huang, Xuanjing  and
      Specia, Lucia  and
      Yih, Scott Wen-tau",
    booktitle = "Proceedings of the 2021 Conference on Empirical Methods in Natural Language Processing",
    month = nov,
    year = "2021",
    address = "Online and Punta Cana, Dominican Republic",
    publisher = "Association for Computational Linguistics",
    url = "https://aclanthology.org/2021.emnlp-main.72/",
    doi = "10.18653/v1/2021.emnlp-main.72",
    pages = "932--948"
}

@inproceedings{gulordava-etal-2018-colorless,
    title = "Colorless Green Recurrent Networks Dream Hierarchically",
    author = "Gulordava, Kristina  and
      Bojanowski, Piotr  and
      Grave, Edouard  and
      Linzen, Tal  and
      Baroni, Marco",
    editor = "Walker, Marilyn  and
      Ji, Heng  and
      Stent, Amanda",
    booktitle = "Proceedings of the 2018 Conference of the North {A}merican Chapter of the Association for Computational Linguistics: Human Language Technologies, Volume 1 (Long Papers)",
    month = jun,
    year = "2018",
    address = "New Orleans, Louisiana",
    publisher = "Association for Computational Linguistics",
    url = "https://aclanthology.org/N18-1108/",
    doi = "10.18653/v1/N18-1108",
    pages = "1195--1205"
}

@article{linzen2016assessing,
  title={Assessing the ability of {LSTMs} to learn syntax-sensitive dependencies},
  author={Linzen, Tal and Dupoux, Emmanuel and Goldberg, Yoav},
  journal={Transactions of the Association for Computational Linguistics},
  volume={4},
  pages={521--535},
  year={2016},
  doi={10.1162/tacl_a_00115},
  publisher={MIT Press}
}

@inproceedings{ettinger2016probing,
  title={Probing for semantic evidence of composition by means of simple classification tasks},
  author={Ettinger, Allyson and Elgohary, Ahmed and Resnik, Philip},
  booktitle={Proceedings of the 1st Workshop on Evaluating Vector-Space Representations for NLP},
  pages={134--139},
  year={2016}
}

@article{miller1995wordnet,
  title={{WordNet: a lexical database for English}},
  author={Miller, George A},
  journal={{Communications of the ACM}},
  volume={38},
  number={11},
  pages={39--41},
  year={1995},
  publisher={ACM New York, NY, USA}
}

@article{misra2022minicons,
    title={minicons: Enabling Flexible Behavioral and Representational Analyses of Transformer Language Models},
    author={Kanishka Misra},
    journal={arXiv:2203.13112},
    year={2022},
    url={https://arxiv.org/abs/2203.13112}
}

@inproceedings{wu-etal-2024-pyvene,
    title = "pyvene: A Library for Understanding and Improving {P}y{T}orch Models via Interventions",
    author = "Wu, Zhengxuan and Geiger, Atticus and Arora, Aryaman and Huang, Jing and Wang, Zheng and Goodman, Noah and Manning, Christopher and Potts, Christopher",
    editor = "Chang, Kai-Wei and Lee, Annie and Rajani, Nazneen",
    booktitle = "Proceedings of the 2024 Conference of the North American Chapter of the Association for Computational Linguistics: Human Language Technologies (Volume 3: System Demonstrations)",
    month = jun,
    year = "2024",
    address = "Mexico City, Mexico",
    publisher = "Association for Computational Linguistics",
    url = "https://aclanthology.org/2024.naacl-demo.16",
    pages = "158--165",
}

@inproceedings{wolf-etal-2020-transformers,
    title = "Transformers: State-of-the-Art Natural Language Processing",
    author = "Wolf, Thomas  and
      Debut, Lysandre  and
      Sanh, Victor  and
      Chaumond, Julien  and
      Delangue, Clement  and
      Moi, Anthony  and
      Cistac, Pierric  and
      Rault, Tim  and
      Louf, Remi  and
      Funtowicz, Morgan  and
      Davison, Joe  and
      Shleifer, Sam  and
      von Platen, Patrick  and
      Ma, Clara  and
      Jernite, Yacine  and
      Plu, Julien  and
      Xu, Canwen  and
      Le Scao, Teven  and
      Gugger, Sylvain  and
      Drame, Mariama  and
      Lhoest, Quentin  and
      Rush, Alexander",
    editor = "Liu, Qun  and
      Schlangen, David",
    booktitle = "Proceedings of the 2020 Conference on Empirical Methods in Natural Language Processing: System Demonstrations",
    month = oct,
    year = "2020",
    address = "Online",
    publisher = "Association for Computational Linguistics",
    url = "https://aclanthology.org/2020.emnlp-demos.6/",
    doi = "10.18653/v1/2020.emnlp-demos.6",
    pages = "38--45"
}

@incollection{rumelhart1993learning,
  title={Learning and connectionist representations},
  author={Rumelhart, David E and Todd, Peter M},
  editor={Meyer, David E and Kornblum, Sylvan},
  booktitle={Attention and Performance {XIV}: Synergies in Experimental Psychology, Artificial Intelligence, and Cognitive Neuroscience},
  pages={3--30},
  year={1993},
  publisher={MIT Press}
}

@article{berko1958child,
  title={The child's learning of {E}nglish morphology},
  author={Berko, Jean},
  journal={Word},
  volume={14},
  number={2-3},
  pages={150--177},
  year={1958},
  publisher={Taylor \& Francis}
}

@inproceedings{jian-manning-2026-humans,
    title = "Humans and transformer {LM}s: Abstraction drives language learning",
    author = "Jian, Jasper  and
      Manning, Christopher D",
    editor = "Demberg, Vera  and
      Inui, Kentaro  and
      Marquez, Llu{\'i}s",
    booktitle = "Proceedings of the 19th Conference of the {E}uropean Chapter of the {A}ssociation for {C}omputational {L}inguistics (Volume 1: Long Papers)",
    month = mar,
    year = "2026",
    address = "Rabat, Morocco",
    publisher = "Association for Computational Linguistics",
    url = "https://aclanthology.org/2026.eacl-long.32/",
    doi = "10.18653/v1/2026.eacl-long.32",
    pages = "752--765",
    ISBN = "979-8-89176-380-7"
}

@inproceedings{oba-etal-2024-language,
    title = "Can Language Models Induce Grammatical Knowledge from Indirect Evidence?",
    author = "Oba, Miyu  and
      Oseki, Yohei  and
      Fukatsu, Akiyo  and
      Haga, Akari  and
      Ouchi, Hiroki  and
      Watanabe, Taro  and
      Sugawara, Saku",
    editor = "Al-Onaizan, Yaser  and
      Bansal, Mohit  and
      Chen, Yun-Nung",
    booktitle = "Proceedings of the 2024 Conference on Empirical Methods in Natural Language Processing",
    month = nov,
    year = "2024",
    address = "Miami, Florida, USA",
    publisher = "Association for Computational Linguistics",
    url = "https://aclanthology.org/2024.emnlp-main.1146/",
    doi = "10.18653/v1/2024.emnlp-main.1146",
    pages = "20591--20603"
}

@article{hohle2004functional,
  title={Functional elements in infants' speech processing: The role of determiners in the syntactic categorization of lexical elements},
  author={H{\"o}hle, Barbara and Weissenborn, J{\"u}rgen and Kiefer, Dorothea and Schulz, Antje and Schmitz, Michaela},
  journal={Infancy},
  volume={5},
  number={3},
  pages={341--353},
  year={2004},
  publisher={Wiley Online Library}
}

@article{lupyan2013difficulties,
  title={The difficulties of executing simple algorithms: Why brains make mistakes computers don’t},
  author={Lupyan, Gary},
  journal={Cognition},
  volume={129},
  number={3},
  pages={615--636},
  year={2013},
  publisher={Elsevier}
}

@article{studdiford2026reasoning,
  title={Reasoning as pattern matching: Shared mechanisms in human and {LLM} everyday reasoning},
  author={Studdiford, Zach and Lupyan, Gary},
  journal={arXiv preprint arXiv:2606.13607},
  year={2026}
}

@article{lampinen2024language,
  title={Language models, like humans, show content effects on reasoning tasks},
  author={Lampinen, Andrew K and Dasgupta, Ishita and Chan, Stephanie CY and Sheahan, Hannah R and Creswell, Antonia and Kumaran, Dharshan and McClelland, James L and Hill, Felix},
  journal={PNAS nexus},
  volume={3},
  number={7},
  pages={pgae233},
  year={2024},
  publisher={Oxford University Press US}
}

@inproceedings{hobbs-mccoy-2026-collocational,
    title = "Collocational bootstrapping: A hypothesis about the learning of subject-verb agreement in humans and neural networks",
    author = "Hobbs, Claire  and
      McCoy, R. Thomas",
    editor = "Bonial, Claire  and
      Berzak, Yevgeni",
    booktitle = "Proceedings of the 30th Conference on Computational Natural Language Learning",
    month = jul,
    year = "2026",
    address = "San Diego, California, USA",
    publisher = "Association for Computational Linguistics",
    url = "https://aclanthology.org/2026.conll-main.7/",
    doi = "10.18653/v1/2026.conll-main.7",
    pages = "90--103",
    ISBN = "979-8-89176-410-1"
}

@article{bock1992regulating,
  title={Regulating mental energy: Performance units in language production},
  author={Bock, Kathryn and Cutting, J Cooper},
  journal={Journal of Memory and Language},
  volume={31},
  number={1},
  pages={99--127},
  year={1992},
  publisher={Elsevier}
}

@article{demuth2020exemplar,
  title   = {Exemplar-based learning probably requires learning abstractions: A commentary on {Ambridge} (2020)},
  author  = {Demuth, Katherine and Johnson, Mark},
  journal = {First Language},
  volume  = {40},
  number  = {5-6},
  pages   = {573--575},
  year    = {2020},
  doi     = {10.1177/0142723720903609}
}

@article{lieven2020similarity,
  title   = {Similarity, analogy and development in radical exemplar theory: A commentary on {Ambridge} (2020)},
  author  = {Lieven, Elena and Ferry, Alissa and Theakston, Anna and Twomey, Katherine E.},
  journal = {First Language},
  volume  = {40},
  number  = {5-6},
  year    = {2020},
  doi     = {10.1177/0142723720909554}
}

@article{franck2002subject,
  title={Subject-verb agreement errors in {F}rench and {E}nglish: The role of syntactic hierarchy},
  author={Franck, Julie and Vigliocco, Gabriella and Nicol, Janet},
  journal={Language and Cognitive Processes},
  volume={17},
  number={4},
  pages={371--404},
  year={2002},
  publisher={Taylor \& Francis}
}

@inproceedings{arehalli2020neural,
  title={Neural language models capture some, but not all, agreement attraction effects},
  author={Arehalli, Suhas and Linzen, Tal},
  booktitle={Proceedings of the Annual Meeting of the Cognitive Science Society},
  volume={42},
  year={2020}
}

@inproceedings{marvin-linzen-2018-targeted,
    title = "Targeted Syntactic Evaluation of Language Models",
    author = "Marvin, Rebecca  and
      Linzen, Tal",
    editor = "Riloff, Ellen  and
      Chiang, David  and
      Hockenmaier, Julia  and
      Tsujii, Jun{'}ichi",
    booktitle = "Proceedings of the 2018 Conference on Empirical Methods in Natural Language Processing",
    month = oct # "-" # nov,
    year = "2018",
    address = "Brussels, Belgium",
    publisher = "Association for Computational Linguistics",
    url = "https://aclanthology.org/D18-1151/",
    doi = "10.18653/v1/D18-1151",
    pages = "1192--1202"
}

@inproceedings{misra-and-kim-2023-catabs,
    title = "{Abstraction via exemplars? A representational case study on lexical category inference in BERT}",
    author = "Misra, Kanishka  and
      Kim, Najoung",
    booktitle = "BUCLD 48: Proceedings of the 48th annual Boston University Conference on Language Development",
    month = nov,
    year = "2023",
    address = "Boston, USA",
url="https://arxiv.org/abs/2312.03708"
}

@book{rogers2004semantic,
  title={Semantic cognition: A parallel distributed processing approach},
  author={Rogers, Timothy T and McClelland, James L},
  year={2004},
  publisher={MIT press}
}

@article{mueller-etal-2026-quest,
    title = "The Quest for the Right Mediator: Surveying Mechanistic Interpretability for {NLP} Through the Lens of Causal Mediation Analysis",
    author = "Mueller, Aaron  and
      Brinkmann, Jannik  and
      Li, Millicent  and
      Marks, Samuel  and
      Pal, Koyena  and
      Prakash, Nikhil  and
      Rager, Can  and
      Sankaranarayanan, Aruna  and
      Sen Sharma, Arnab  and
      Sun, Jiuding  and
      Todd, Eric  and
      Bau, David  and
      Belinkov, Yonatan",
    journal = "Computational Linguistics",
    volume = "52",
    number = "1",
    month = mar,
    year = "2026",
    address = "Cambridge, MA",
    publisher = "MIT Press",
    url = "https://aclanthology.org/2026.cl-1.10/",
    doi = "10.1162/coli.a.572",
    pages = "331--378"
}

@article{grand2022semantic,
  title={{Semantic projection recovers rich human knowledge of multiple object features from word embeddings}},
  author={Grand, Gabriel and Blank, Idan Asher and Pereira, Francisco and Fedorenko, Evelina},
  journal={Nature Human Behaviour},
  pages={1--13},
  year={2022},
  publisher={Nature Publishing Group},
  url={https://www.nature.com/articles/s41562-022-01316-8}
}

@inproceedings{xu-etal-2026-cross,
    title = "Cross-Modal Taxonomic Generalization in (Vision-) Language Models",
    author = "Xu, Tianyang  and
      Sandoval-Casta{\~n}eda, Marcelo  and
      Livescu, Karen  and
      Shakhnarovich, Greg  and
      Misra, Kanishka",
    editor = "Liakata, Maria  and
      Moreira, Viviane P.  and
      Zhang, Jiajun  and
      Jurgens, David",
    booktitle = "Proceedings of the 64th Annual Meeting of the {A}ssociation for {C}omputational {L}inguistics (Volume 1: Long Papers)",
    month = jul,
    year = "2026",
    address = "San Diego, California, United States",
    publisher = "Association for Computational Linguistics",
    url = "https://aclanthology.org/2026.acl-long.742/",
    doi = "10.18653/v1/2026.acl-long.742",
    pages = "16319--16337",
    ISBN = "979-8-89176-390-6"
}

@inproceedings{geiger2024finding,
  title={Finding alignments between interpretable causal variables and distributed neural representations},
  author={Geiger, Atticus and Wu, Zhengxuan and Potts, Christopher and Icard, Thomas and Goodman, Noah},
  booktitle={Causal Learning and Reasoning},
  pages={160--187},
  year={2024},
  organization={PMLR}
}

@article{jafari2025relp,
  title={{RelP}: Faithful and efficient circuit discovery in language models via relevance patching},
  author={Jafari, Farnoush Rezaei and Eberle, Oliver and Khakzar, Ashkan and Nanda, Neel},
  journal={arXiv preprint arXiv:2508.21258},
  year={2025}
}

@misc{bai2025qwen3,
      title={{Qwen3-VL Technical Report}}, 
      author={Shuai Bai and Yuxuan Cai and Ruizhe Chen and Keqin Chen and Xionghui Chen and Zesen Cheng and Lianghao Deng and Wei Ding and Chang Gao and Chunjiang Ge and Wenbin Ge and Zhifang Guo and Qidong Huang and Jie Huang and Fei Huang and Binyuan Hui and Shutong Jiang and Zhaohai Li and Mingsheng Li and Mei Li and Kaixin Li and Zicheng Lin and Junyang Lin and Xuejing Liu and Jiawei Liu and Chenglong Liu and Yang Liu and Dayiheng Liu and Shixuan Liu and Dunjie Lu and Ruilin Luo and Chenxu Lv and Rui Men and Lingchen Meng and Xuancheng Ren and Xingzhang Ren and Sibo Song and Yuchong Sun and Jun Tang and Jianhong Tu and Jianqiang Wan and Peng Wang and Pengfei Wang and Qiuyue Wang and Yuxuan Wang and Tianbao Xie and Yiheng Xu and Haiyang Xu and Jin Xu and Zhibo Yang and Mingkun Yang and Jianxin Yang and An Yang and Bowen Yu and Fei Zhang and Hang Zhang and Xi Zhang and Bo Zheng and Humen Zhong and Jingren Zhou and Fan Zhou and Jing Zhou and Yuanzhi Zhu and Ke Zhu},
      year={2025},
      eprint={2511.21631},
      archivePrefix={arXiv},
      primaryClass={cs.CV},
      url={https://arxiv.org/abs/2511.21631}, 
}

@article{Dubova_Sloman_2026, title={Excess Capacity Learning}, DOI={10.1017/S0140525X2610510X}, journal={Behavioral and Brain Sciences}, author={Dubova, Marina and Sloman, Sabina J.}, year={2026}, pages={1–77}}

@misc{nanda2023attribution,
  title={Attribution patching: Activation patching at industrial scale},
  author={Nanda, Neel},
  year={2023},
  howpublished={\url{https://www.neelnanda.io/mechanistic-interpretability/attribution-patching}}
}

@inproceedings{
alain2016understanding,
title={Understanding intermediate layers using linear classifier probes},
author={Guillaume Alain and Yoshua Bengio},
year={2017},
booktitle={The Fifth International Conference on Learning Representations},
url={https://openreview.net/forum?id=ryF7rTqgl}
}

@article{marks2023geometry,
  title={The geometry of truth: Emergent linear structure in large language model representations of true/false datasets},
  author={Marks, Samuel and Tegmark, Max},
  journal={arXiv preprint arXiv:2310.06824},
  year={2023}
}

\appendix
\label{sec:appendix}
\raggedbottom

\section{Additional Details of Embeddings Training}
\label{app:embed-train-methods}

\subsection{Embeddings initialization}
\label{app:embed-init}
We initialize all \texttt{[wug]} and \texttt{[wugs]} embeddings instances at the mean of an equal number of real singular and plural nouns ($n=28$ pairs). Because the set of singular and plural embeddings used to extract the mean is evenly balanced, there is no information in this embedding biasing learning towards singular or plural nouns. Table \ref{tab:noun-init} shows all (single token) nouns used for initialization.

\begin{table}[h]
\centering
\small
\begin{tabular}{@{}llllll@{}}
\toprule
cat & cats & dog & dogs & bird & birds \\
bear & bears & rat & rats & tree & trees \\
word & words & thing & things & car & cars \\
house & houses & rock & rocks & chair & chairs \\
table & tables & cup & cups & book & books \\
phone & phones & man & men & woman & women \\
child & children & person & people & door & doors \\
gate & gates & fence & fences & pond & ponds \\
lamp & lamps & corner & corners & path & paths \\
wall & walls &  &  &  &  \\
\bottomrule
\end{tabular}
\caption{The 28 noun pairs used for embeddings initialization.}
\label{tab:noun-init}
\end{table}

\subsection{Vision condition}
\label{app:vis-condition}
\paragraph{Image stimuli} 
Our image stimuli consist of five images of one \texttt{[wug]} creature (generated using the OpenAI API), and five images of two \texttt{[wugs]} creatures. The full set of images can be seen in Figure \ref{fig:wug_stim} below.

\begin{figure}[h]
    \centering
    \includegraphics[width=0.55\linewidth]{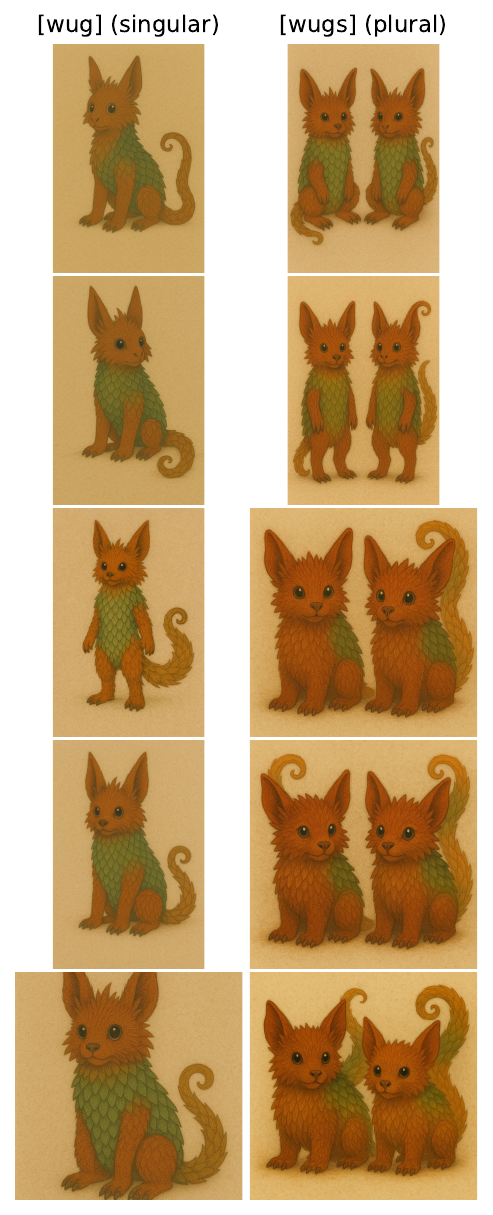}
    \caption{Image stimuli for \texttt{[wug]} and \texttt{[wugs]} training in the vision condition.}
    \label{fig:wug_stim}
\end{figure}

\paragraph{Image stimuli generation prompts} 

We additionally include the prompts used to generate images used for training \texttt{[wug]} and \texttt{[wugs]} embeddings. These queries were used to generate five singular and plural images from the LLM \texttt{openai-o3} via the OpenAI online API. Because the actual word \textit{wug} \citep{berko1958child} and \textit{wug} illustrations are very likely part of large model pretraining corpora, we query \texttt{openai-o3} to generate images of one or multiple \textit{``snarples''} (a name made up by one of the authors), as seen below: 

\nolinenumbers
\begin{lstlisting}[caption={Image-generation prompts used to create singular ``snarple'' stimuli},label={lst:snarple-prompts},basicstyle=\small\ttfamily,frame=single,
columns=fullflexible,breaklines=true]

generate an image of an imaginary creature called a snarple

now generate an image of the snarple facing a different direction

awesome! Now generate it facing the other direction

now facing forward again

now generate an image of singular snarple
\end{lstlisting}

\nolinenumbers
\begin{lstlisting}[
  caption={Image-generation prompts used to create plural ``snarple'' stimuli},
  label={lst:snarple-prompts-pl},
  basicstyle=\small\ttfamily,
  frame=single,
  columns=fullflexible,
  breaklines=true]

now generate an image of two snarples

generate two more images of two snarples each

now do another image of two snarples

now another and they are standing up

awesome now one more sitting again
\end{lstlisting}

\paragraph{Text stimuli} Text stimuli in the vision condition were intentionally designed not to include any discriminating syntactic information, and used the same exact template for both words, so as to only have the discriminating information in the images.
Table \ref{tab:stimuli-image} shows the stimuli templates used for \texttt{[wug]} and \texttt{[wugs]}.

% \begin{table}[H]
% \centering
% \small
% \begin{tabular}{@{}p{0.44\linewidth}p{0.44\linewidth}@{}}
% \toprule
% Singular ({[}wug{]}) & Plural ({[}wugs{]}) \\
% \midrule
% {[}wug{]}? & {[}wugs{]}. \\
% {[}wug{]}, there. & {[}wugs{]}! \\
% The {[}wug{]} from over there. & {[}wugs{]} here. \\
% {[}wug{]} over there. & The {[}wugs{]} from over here. \\
% {[}wug{]} near the fence. & {[}wugs{]} by the rock. \\
% {[}wug{]} around the chair. & {[}wugs{]} behind the tree. \\
% Do you see the {[}wug{]} near the fence? & Look at the {[}wugs{]} by the rock. \\
% I walked past the {[}wug{]} along the path. & I noticed the {[}wugs{]} behind the tree. \\
% I found the {[}wug{]} under the lamp. & I pointed to the {[}wugs{]} in the corner. \\
% I stood beside the {[}wug{]} by the gate. & I watched the {[}wugs{]} from a distance. \\
% I looked for the {[}wug{]} around the chair. & I moved toward the {[}wugs{]} near the pond. \\
% I noticed the {[}wug{]} in the corner by the lamp. & Look at the {[}wugs{]} near the rock by the fence. \\
% I pointed to the {[}wug{]} behind the chair near the wall. & I walked past the {[}wugs{]} near the gate by the path. \\
% I found the {[}wug{]} near the pond by the tree. & I watched the {[}wugs{]} under the lamp by the door. \\
% Do you see the {[}wug{]} by the fence near the gate? & I moved toward the {[}wugs{]} around the chair by the rock. \\
% \bottomrule
% \end{tabular}
% \caption{Training stimuli for the image condition. Each sentence is paired with a singular or plural creature image.}
% \label{tab:stimuli-image}
% \end{table}

\begin{table}[h]
\centering
\small
\begin{tabular}{@{}p{0.44\linewidth}p{0.44\linewidth}@{}}
\toprule
Singular (\wug{}) & Plural (\wugs{}) \\
\midrule
Look at the \wug{}. & Look at the \wugs{}. \\
I spotted the \wug{} in the picture. & I spotted the \wugs{} in the picture. \\
I noticed the \wug{} on the screen. & I noticed the \wugs{} on the screen. \\
A picture was taken of the \wug{}. & A picture was taken of the \wugs{}. \\
I drew the \wug{} on the page. & I drew the \wugs{} on the page. \\
They pointed at the \wug{} on the screen. & They pointed at the \wugs{} on the screen. \\
We looked closely at the \wug{}. & We looked closely at the \wugs{}. \\
I circled the \wug{} in the image. & I circled the \wugs{} in the image. \\
Someone highlighted the \wug{} in the picture. & Someone highlighted the \wugs{} in the picture. \\
I found the \wug{} in this image. & I found the \wugs{} in this image. \\
We noticed the \wug{} right away. & We noticed the \wugs{} right away. \\
I took a screenshot of the \wug{}. & I took a screenshot of the \wugs{}. \\
They zoomed in on the \wug{}. & They zoomed in on the \wugs{}. \\
I stared at the \wug{} for a while. & I stared at the \wugs{} for a while. \\
I glanced at the \wug{} in the photo. & I glanced at the \wugs{} in the photo. \\
We studied the \wug{} closely. & We studied the \wugs{} closely. \\
\bottomrule
\end{tabular}
\caption{Training stimuli for the image condition. The singular and plural captions are identical except for the novel token. Each caption is paired with a singular or plural creature image.}
\label{tab:stimuli-image}
\end{table}

% \pagebreak
\subsection{Language condition}
\label{app:lang-condition}

\paragraph{Text stimuli}
Text stimuli ($n=30$) in the language condition were designed to include syntactic cues informative of whether \texttt{[wug]} and \texttt{[wugs]} embeddings corresponded to singular or plural nouns. All stimuli were verified prior to inclusion.

\begin{table}[h]
\centering
\small
\begin{tabular}{@{}p{0.44\linewidth}p{0.44\linewidth}@{}}
\toprule
Singular ({[}wug{]}) & Plural ({[}wugs{]}) \\
\midrule
One {[}wug{]}. & Several {[}wugs{]}. \\
A single {[}wug{]}. & A few {[}wugs{]}. \\
There sits a {[}wug{]}. & There sit some {[}wugs{]}. \\
I spotted a {[}wug{]}. & I spotted some {[}wugs{]}. \\
That is definitely a {[}wug{]}. & Those are definitely {[}wugs{]}. \\
The {[}wug{]} shook its tail before running away. & The {[}wugs{]} shook their tails before running away. \\
Beneath the stairs crouched a frightened {[}wug{]}. & Beneath the stairs crouched several frightened {[}wugs{]}. \\
Does the {[}wug{]} always come back at night? & Do the {[}wugs{]} always come back at night? \\
One {[}wug{]} carries a small leaf in its mouth. & Many {[}wugs{]} carry small leaves in their mouths. \\
The {[}wug{]} near the window makes a soft sound. & The {[}wugs{]} near the window make a soft sound. \\
A {[}wug{]} appeared from behind the curtain. & A dozen {[}wugs{]} appeared from behind the curtain. \\
That {[}wug{]} seems friendlier than the others. & Those {[}wugs{]} seem friendlier than the others. \\
Along the path I discovered a lone {[}wug{]}. & Along the path I discovered a group of {[}wugs{]}. \\
The {[}wug{]} was climbing the hill before sunset. & The {[}wugs{]} were climbing the hill before sunset. \\
Curled up in the corner was a sleeping {[}wug{]}. & Curled up in the corner were several sleeping {[}wugs{]}. \\
\bottomrule
\end{tabular}
\caption{Training stimuli for the language condition.}
\label{tab:stimuli-syntax}
\end{table}

\subsection{Dev-Set Sentence Construction}
\label{app:dev-set-construct}
As a criteria for halting embeddings training after some number of epochs, for the language condition, we create a \textit{``dev-set''} consisting of $n=280$ minimal pairs sentences evaluating number agreement in \texttt{[wug]} and \texttt{[wugs]} embeddings across a range of constructions. We first manually created a limited set of constructions, and then augmented these further using a frontier LLM (\texttt{openai-o3}). All stimuli were then manually verified. \Cref{tab:dev-language} shows some examples.

\begin{table}[h]
\centering
\small
\begin{tabular}{@{}p{0.44\linewidth}p{0.44\linewidth}@{}}
\toprule
Singular & Plural \\
\midrule
The \wug{}/*\wugs{} is glowing faintly under the lamp. & The \wugs{}/*\wug{} have been sleeping all afternoon. \\
Every \wug{}/*\wugs{} needs a safe place to rest. & After the two \wugs{}/*\wug{} left, the garden felt empty. \\
There is a \wug{}/*\wugs{} sitting on the roof. & The \wugs{}/*\wug{} that escaped were found near the stream. \\
The \wug{}/*\wugs{} they chased is quick. & The \wugs{}/*\wug{} chasing birds are fast. \\
Across the field runs a \wug{}/*\wugs{}. & Under the bridge live many \wugs{}/*\wug{}. \\
\bottomrule
\end{tabular}
\caption{Examples of development stimuli for the language condition. Each item is a minimal pair that differs only in the novel noun, and the model is expected to prefer the grammatical noun over the ungrammatical one (marked with *).}
\label{tab:dev-language}
\end{table}

% \pagebreak
% \pagebreak

\subsection{Instruction Template Format}
\label{app:full-instruct-template}
Because \texttt{Qwen-3-VL-*} models post-training included a chat template, we also interpolate our prompts in chat templates both for training and evaluation. We use the following chat template for training in the vision condition:

\nolinenumbers
\begin{lstlisting}[caption={Chat template for the image condition.},
label={lst:template-image},basicstyle=\small\ttfamily,frame=single,
columns=fullflexible,breaklines=true]
<|im_start|>user
<|vision_start|><|image_pad|><|vision_end|>Caption this image.<|im_end|>
<|im_start|>assistant
Look at the [wug].<|im_end|>
\end{lstlisting}

And the following chat template in all other cases (language condition, evaluations):

\nolinenumbers
\begin{lstlisting}[caption={Chat template for the language condition and for all evaluations.},
label={lst:template-syntax},basicstyle=\small\ttfamily,frame=single,
columns=fullflexible,breaklines=true]
<|im_start|>user
Complete the sentence.<|im_end|>
<|im_start|>assistant
One [wug].<|im_end|>
\end{lstlisting}

\section{Additional Results of Embeddings Training}
\label{app:embed-train-results}

\subsection{LR hyperparameter sweep}
\label{app:hyperparam-sweep}
\begin{figure}[h]
    \centering
    \includegraphics[width=0.8\linewidth]{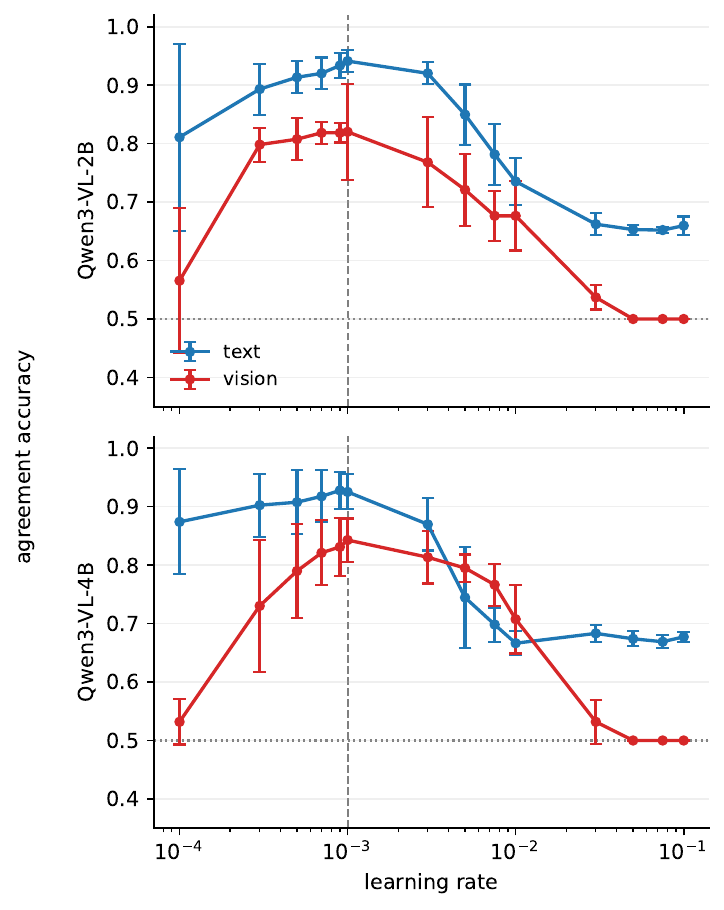}
    \caption{Results of LR sweep for all models and learning conditions.}
    \label{fig:lr_sweep}
\end{figure}

We systematically sweep across a range of learning rates for language and vision conditions in both models (training at five random seed initializations at each learning rate evaluated). We find that $10^{-3}$ is the optimal learning rate for both models and conditions, and use this learning rate for all subsequent embeddings training. Figure \ref{fig:lr_sweep} shows the results of the full LR sweep.

\subsection{Loss curves for learned embeddings} 
\label{app:loss-curves}
\begin{figure}[h]
    \centering
    \includegraphics[width=0.8\linewidth]{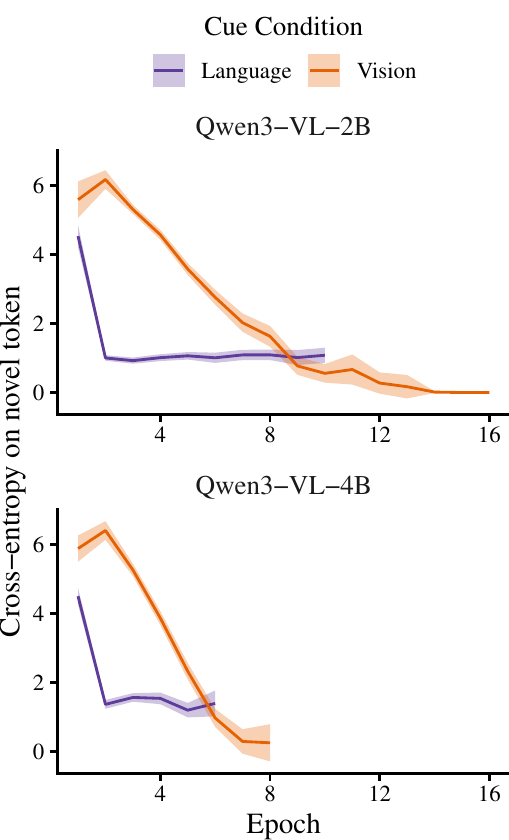}
    \caption{Cross-entropy loss for \texttt{[wug]} and \texttt{[wugs]} training, for 50 seeds in each condition. Because of early stopping, runs end at different
epochs; we show epochs in which at least 10 seeds are still training.}
    \label{fig:wug_loss}
\end{figure}

We include loss curves for the mean cross-entropy loss of \texttt{[wug]} and \texttt{[wugs]} tokens in the vision and language conditions for both models, for sentences seen in training. Loss curves are averaged across the full set of $n=50$ runs for different seed initializations at the best performing learning rate. These loss trajectories can be seen in Figure \ref{fig:wug_loss}. Interestingly, we note greater within-distribution loss for the language (text based) condition as opposed to vision.

\subsection{Examples of productive generation of novel embeddings}
\label{app:embed-generate}

The bulk of our main results compare relative log probabilities between \texttt{[wug]} and \texttt{[wugs]}. However, are models able to actually generate the novel learned embeddings when appropriate? We use \texttt{minicons} library \citep{misra2022minicons} to generate next token completions for several base sentences contain \texttt{[wug]} or \texttt{[wugs]} tokens following embeddings training in \texttt{Qwen-3-VL-2B-Instruct}. Qualitatively, we note that the model is indeed able to productively generate the next token: that is, the relative probabilities for \texttt{[wug]} and \texttt{[wugs]} are not buried in the greater logit distribution (nor does the model deterministically generate only \texttt{[wug]} or \texttt{[wugs]}), at least for these examples. Table \ref{tab:freeform-generations} shows base sentences and completions for \texttt{Qwen-3-VL-2B-Instruct}. One exception to this trend was stimulus number 3, where the model correctly generate \wug{} for the input ``\textit{I drew a \wug{} on the page. Then I drew \textbf{one} more}'', but it just ends the sentence with a period for ``\textit{I drew a \wug{} on the page. Then I drew \textbf{four} more}''.

% \begin{table}[h]
% \centering
% \small
% \begin{tabular}{@{}rp{0.42\linewidth}p{0.42\linewidth}@{}}
% \toprule
% \# & Prompt & Model generation \\
% \midrule
% 1 & One {[}wug{]} was playing and another came to join it. Now there are
%   & One {[}wug{]} was playing and another came to join it. Now there are \textbf{{[}wugs{]}} \\
% \addlinespace
% 2 & I saw a single {[}wug{]} yesterday. Today I saw three
%   & I saw a single {[}wug{]} yesterday. Today I saw three \textbf{{[}wugs{]}} \\
% \addlinespace
% 3 & There is one {[}wug{]} on the left and two
%   & There is one {[}wug{]} on the left and two \textbf{{[}wugs{]} on the right.} \\
% \addlinespace
% 4 & I saw one {[}wug{]}. Then another joined. now there are
%   & I saw one {[}wug{]}. Then another joined. now there are \textbf{{[}wugs{]}} \\
% \addlinespace
% 5 & If a {[}wug{]} is lonely then it should find {[}wug{]}
%   & If a {[}wug{]} is lonely then it should find \textbf{{[}wugs{]}} \\
%   \addlinespace
% 6 & There is one {[}wug{]} on the left and one
%   & There is one {[}wug{]} on the left and one \textbf{{[}wug{]} on the right.} \\
% \bottomrule
% \end{tabular}
% \caption{Generations from Qwen3-VL-2B with the learned \wug{}/\wugs{}
% embeddings. Generated tokens are in \textbf{bold}}
% \label{tab:freeform-generations}
% \end{table}

\begin{table}[t]
\centering
\small
\begin{tabular}{@{}lp{0.55\linewidth}p{0.3\linewidth}@{}}
\toprule
\# & Prompt & Generation \\
\midrule
1 & There is one \wug{} on the left and \{one / two\} small & \textbf{\wug{}} / \textbf{\wugs{}} on the right. \\
2 & Yesterday I fed one \wug{}. Today I fed \{one / three\} hungry & \textbf{\wug{}} / \textbf{\wugs{}}. \\
3 & I drew a \wug{} on the page. Then I drew \{one / four\} more & \textbf{\wug{}} / (ends sentence) \\
4 & We saw a \wug{} at the park. Later we saw \{a / several\} sleepy & \textbf{\wug{}} / \textbf{\wugs{}}. \\
5 & There are two \wugs{} on the left and \{one / two\} small & \textbf{\wug{}} / \textbf{\wugs{}} on the right. \\
6 & Yesterday I fed three \wugs{}. Today I fed \{one / three\} hungry & \textbf{\wug{}} / \textbf{\wugs{}}. \\
7 & I drew some \wugs{} on the page. Then I drew \{one / four\} more & \textbf{\wug{}} / \textbf{\wugs{}}. \\
8 & We saw many \wugs{} at the park. Later we saw \{a / several\} sleepy & \textbf{\wug{}} / \textbf{\wugs{}}. \\
\bottomrule
\end{tabular}
\caption{Greedy generations from Qwen3-VL-2B with the learned \wug{}/\wugs{} embeddings (vision condition, best seed), for pairs of prompts that differ only in the quantity word. Generations for the singular and plural prompt are separated by ``/''.}
\label{tab:freeform-generations}
\end{table}

\subsection{Effect of image choice}
\label{sec:img-choice}

One concern with our vision cue condition is that all of our images show the same creature (which we call \textit{snarple} here), so our results could be specific to this one creature. To test this, we trained the vision-cued embeddings for Qwen3-VL-2B on two other kinds of images: a different creature (the warbles), and a simple pink blob on a white background, where plural images show 2--4 blobs. Everything else is the same as in our main experiments: we use the same captions, select the learning rate in the same way using held-out images of each object and train 50 seeds for each.

\begin{figure*}[t]
    \centering
    \includegraphics[width=0.8\textwidth]{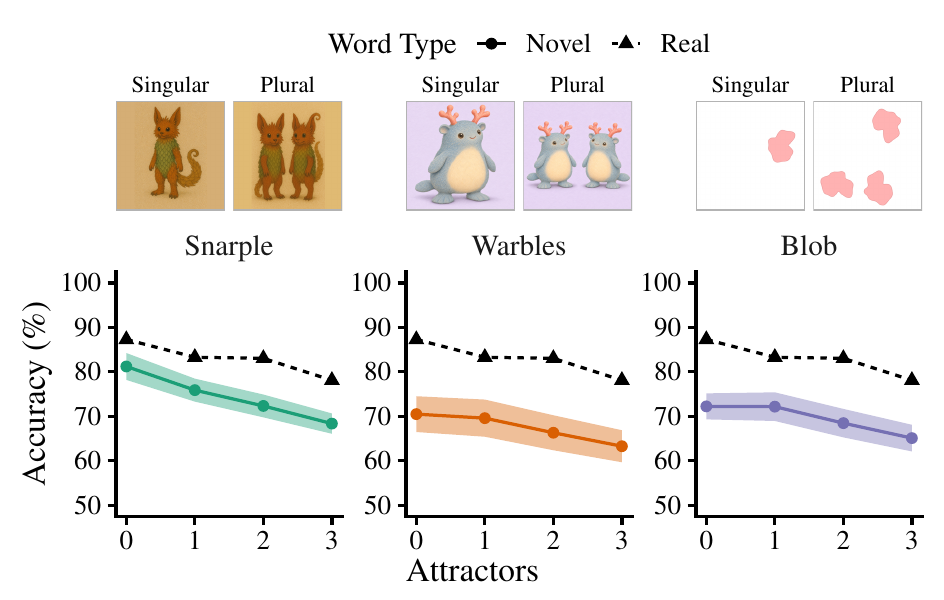}
    \caption{Number agreement accuracy (\%) of Qwen3-VL-2B on novel words learned from three different referents (top: example singular and plural training images), compared to real nouns. Shaded regions show 95\% confidence intervals across 50 seeds.}
    \label{fig:objects}
\end{figure*}

\Cref{fig:objects} shows our results. Similar to the main text results, we see that models are above chance at number agreement for all attractors, regardless of the images used to learn the novel nouns. It is worth noting that the accuracy is lower for the warbles (67.4\%) and the blobs (69.5\%) than for the \textit{snarple} (74.4\%), and this difference is largest when there are no attractors.

\section{Results on Qwen3-VL-4B}
\label{app:2b-results}

\subsection{Replication of embeddings movement during training}
\label{app:embed-move-2b}
\begin{figure}[h]
    \centering
    \includegraphics[width=0.8\linewidth]{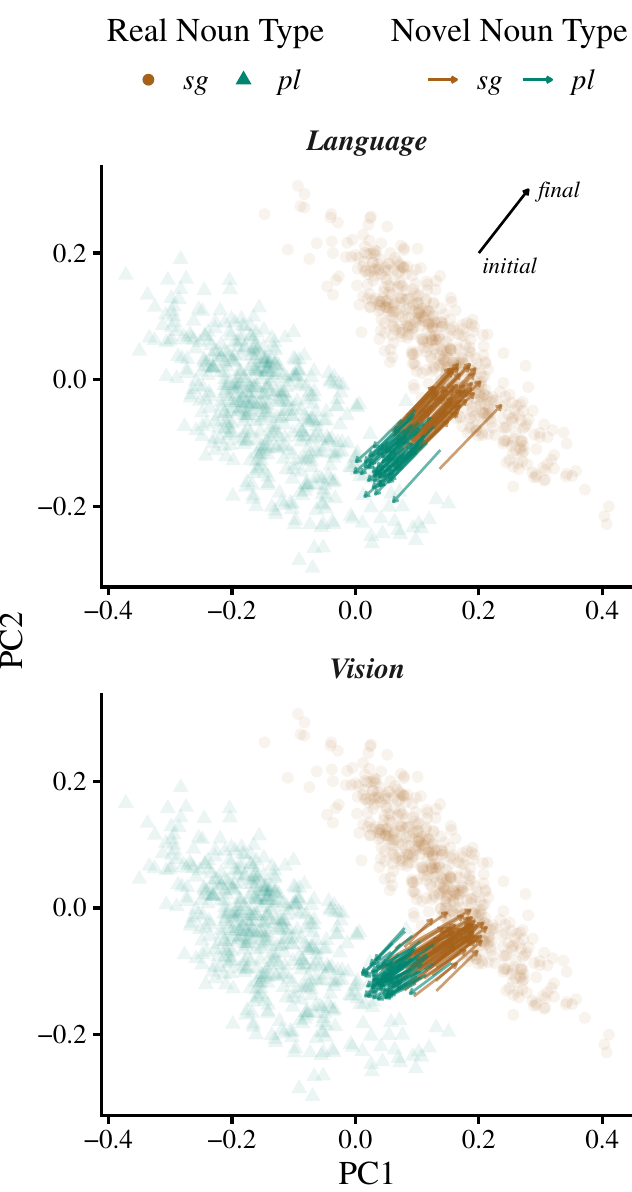}
    \caption{Movement (shown using arrows) of the embedding states of novel words when analyzed using a 2D PCA fit on embeddings of real singular (\textit{sg}) and plural (\textit{pl}) nouns (e.g., \textit{dogs, chairs, blocks,} etc., $N$=500 each) in the Qwen3-VL-4B model across both cue conditions (Language and Vision). Colors indicate grammatical number for real and novel nouns.}
    % \label{fig:movement-pca-2b}
    \label{fig:movement-pca-4b}
\end{figure}

We replicate the results shown in Figure \ref{fig:movement-pca-2b} for Qwen-3-VL-4B as shown in \Cref{fig:movement-pca-4b}, observing qualitatively similar results. Both \texttt{[wug]} and \texttt{[wugs]} nouns representations diverge along trajectories defined by real singular and plural nouns.

\subsection{Mechanistic evidence for Qwen3-VL-4B}
\label{app:mech4b}

\begin{figure*}[!ht]
    \centering
    \includegraphics[width=0.8\textwidth]{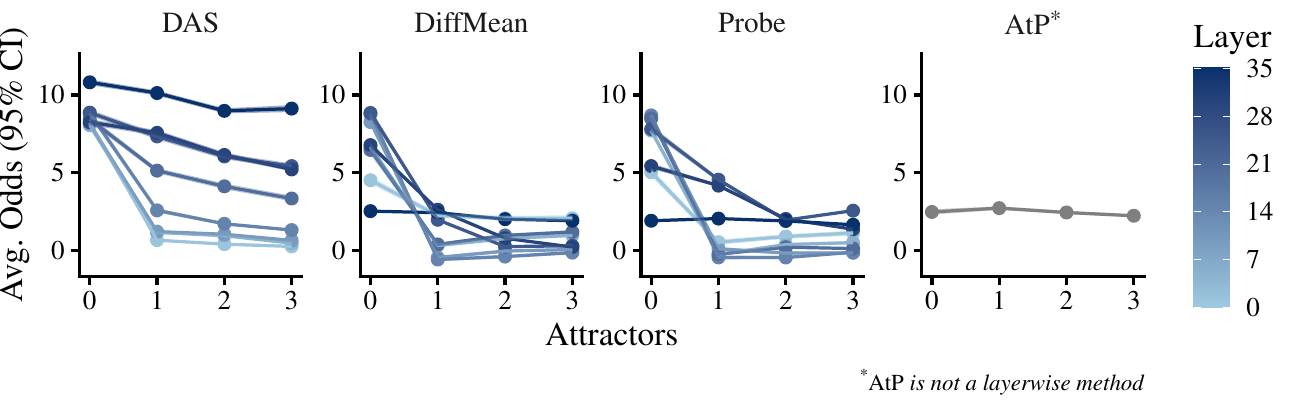}
    \caption{Avg. Odds across model layers (when applicable) for the Qwen3-VL-4B model on agreement stimuli with real nouns as subjects. Higher values mean greater causal effect.}
    \label{fig:real-interp-4b}
\end{figure*}

\begin{figure*}[!ht]
    \centering
    \includegraphics[width=0.8\linewidth]{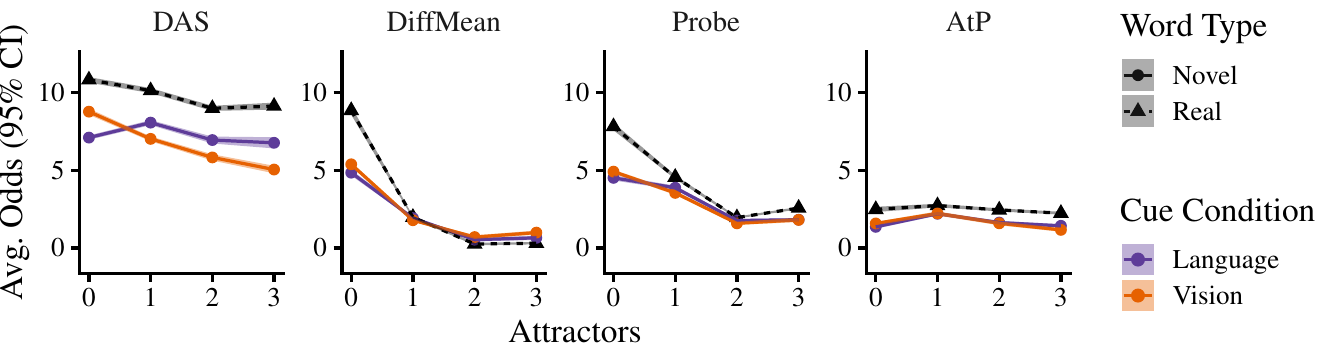}
    \caption{Avg. Odds for the Qwen3-VL-4B model on agreement stimuli with subjects that are real vs. novel nouns acquired from both types of Cue conditions. Results shown for layer with best overall avg. odds chosen on results on stimuli with real nouns as subjects. We see generally high agreement in the results across both cue-conditions.}
    \label{fig:novel-vs-real-interp-4b}
    \vspace{-1em}
\end{figure*}

Results on performing mechanistic analyses for Qwen3-VL-4B are shown in \cref{fig:real-interp-4b} and \cref{fig:novel-vs-real-interp-4b}. We see generally similar results as in the 2B model, albeit with slightly weaker avg. odds in the novel word results. Nevertheless, we see further evidence of no difference between language vs. vision cue conditions.

\section{Details of Intervention Methods}
\label{app:intervention-details}

\paragraph{Distributed Alignment Search (DAS) \citep{arora-etal-2024-causalgym}}: DAS learns a 1-dimensional subspace in model activations with the objective of maximizing the likelihood of a given counterfactual completion token for some input sequence. Specifically, given a \textit{source input} $\mathbf{s}$ and a \textit{base input} $\mathbf{b}$ we wish to intervene on, DAS learns the rotation:
\begin{equation}
    h' = h_{\mathbf{b}} + \big( h_{\mathbf{s}} \mathbf{a}^\top - h_{\mathbf{b}} \mathbf{a}^\top \big)\, \mathbf{a},
    \label{eq:das}
\end{equation}
where $h_{\mathbf{b}}, h_{\mathbf{s}} \in \mathbb{R}^{1 \times d}$ are the residual stream activations at layer $\ell$ and some token position $p$ for the base and source inputs respectively, and $\mathbf{a} \in \mathbb{R}^{1 \times d}$ is the learned direction. In our case, $\mathbf{s}$ and $\mathbf{b}$ are a minimal-difference singular--plural sentence pair $\langle s_s, s_p \rangle$, where $\mathbf{a}$ is learned to maximize the relative likelihood of the counterfactual completion, $\log p(s_s^{\text{cf}}) > \log p(s_s^{\text{base}})$. The explicit supervised learning signal in DAS (which learns the rotation $h'$ over many labeled pairs) guarantees that the method will converge to the 1-dimensional subspace which captures the difference in model behavior between \textbf{s} and \textbf{b} given one exists. 

\paragraph{Linear Probe \citep{alain2016understanding, ettinger2016probing}:} Given the set of residual stream activations for singular and plural sentences $\langle s_s, s_p \rangle$ at a given layer $\ell$, we apply a logistic regression to classify activations from labeled singular ($y = 0$) and plural ($y = 1$) sentences. Given the regression for classifying singular and plural sentences:
\begin{equation}
    p(y = 1 \mid h) = \sigma(h \mathbf{w}^\top)
    \label{eq:probe}
\end{equation}
we intervene on model representations using the weight vector $\mathbf{w} \in \mathbb{R}^{1 \times d}$, adding either $+\alpha \hat{\mathbf{w}}$ to the residual stream to increase the likelihood of the plural completion or $-\alpha \hat{\mathbf{w}}$ to increase the likelihood of the singular completion, where $\alpha$ is a steering coefficient.
    
\paragraph{Difference-of-Means (DiffMean) \citep{marks2023geometry}:} Given the set of activations for singular and plural sentences $\langle \mathcal{H}_s, \mathcal{H}_p \rangle$ at layer $\ell$, difference-in-means uses the vector
\begin{equation}
    \mathbf{v} = \mu_p - \mu_s
    \label{eq:diffmean}
\end{equation}
as the intervention on a given minimal-difference pair $\langle s_s, s_p \rangle$, adding $+\alpha \hat{\mathbf{v}}$ to increase the likelihood of the plural completion and $-\alpha \hat{\mathbf{v}}$ for the singular.
    
\paragraph{Attribution patching \citep{nanda2023attribution}:} Rather than finding a subspace in the residual stream, Attribution patching first estimates the set of individual model components with the greatest estimated effect on the target behavior via a gradient based approximation of activation patching, retaining the top-$k$ parameters based on the attribution value:
\begin{equation}
\hat{c}(n) = \mathop{\mathbb{E}}_{(\mathbf{b},\mathbf{s})}\big[(n(\mathbf{s}) - n(\mathbf{b}))^{\top} \nabla_{n}\mathcal{L}(\mathbf{b})\big].
\end{equation}

where $n(\mathbf{b}), n(\mathbf{s})$ are the activations of component $n$ under the base and source inputs, and $\nabla_{n}\mathcal{L}(\mathbf{b})$ is the gradient of the difference in logits between the completion agreeing with the source sentence and the completion agreeing with the base sentence $\mathcal{L} = \mathrm{logit}(v_{\mathbf{s}}) - \mathrm{logit}(v_{\mathbf{b}})$ (e.g.\ $\mathrm{logit}(\textit{are}) - \mathrm{logit}(\textit{is})$). We then patch the activations for the top-$k$ model components between base and source sentences. While attribution patching can in principle be applied to any set of model components, we focus on MLP intermediate activations (which we find obtain the best performance on natural sentences, see Appendix \ref{app:atp-k-select}).    

%
% where $n(\mathbf{b}), n(\mathbf{s})$ are the activations of parameter $n$ under the base and source inputs, and $\rho_n$ is the relevance score propagated backward from the output to $n$, evaluated at $n(\mathbf{b})$. We then interchange the activations of those components directly between minimal-difference (source, base) sentence pairs, where $\mathcal{C}$ is the top-$k$ components by $|\hat{c}(n)|$:
% %
% \begin{equation}
%     \mathcal{M}\big( \mathbf{b} \mid (n \leftarrow n(\mathbf{s})) \ \ \forall\, n \in \mathcal{C} \big).
%     \label{eq:relp-patch}
% \end{equation}

% Appendix \ref{} includes additional details on our justification for selecting a specific number of parameters $k$ and activation basis (in our case, MLP intermediate activations).

\section{Statistical Analyses for Cross-modal Mechanistic Analyses}
\label{sec:stats}

To understand the effect of our various considerations in the mechanistic analyses, we conduct a statistical test using linear mixed effects regression model which we fit on results from both models, across all attractors, and all layerwise methods. We predict avg. odds using cue condition, attractors, model, and interpretability method as fixed effects, and layer as random effects:
\begin{align*}
    \texttt{AvgOdds} &\sim \texttt{cue} + \texttt{attractors} + \texttt{method} \\&+ \texttt{model} + (1 \mid \texttt{layer})
\end{align*}

Results from this analysis are shown in \Cref{tab:lmm-avgodds}. We see that except for the cue condition, all other predictors show significant effects. For instance, both DiffMean and Probe have worse odds than DAS (as clearly seen in our results), while odds decrease with increase in the number of attractors.

% \begin{table*}[!ht]
% \centering
% \begin{tabular}{lrrrrr}
% \toprule
% Term & $\beta$ & SE & \textit{df} & $t$ & $p$ \\
% \midrule
% (Intercept)                            & 4.613  & 0.375 & 13.0  & 12.31  & \textless 0.001 \\
% \texttt{cue}$_{\textit{Vision}}$       & $-$0.016 & 0.157 & 322.7 & $-$0.10  & 0.920 \\
% \texttt{attractors}                    & $-$0.930 & 0.070 & 322.7 & $-$13.25 & \textless 0.001 \\
% \texttt{method}$_{\textit{DiffMean}}$  & $-$1.861 & 0.192 & 322.7 & $-$9.68  & \textless 0.001 \\
% \texttt{method}$_{\textit{Probe}}$     & $-$1.895 & 0.192 & 322.7 & $-$9.86  & \textless 0.001 \\
% \texttt{model}$_{\textit{Qwen3-VL-4B}}$& 0.313  & 0.168 & 328.8 & 1.86   & 0.064 \\
% \bottomrule
% \end{tabular}
% \caption{Linear mixed-effects model results}
% \label{tab:lmm-avgodds}
% \end{table*}

\begin{table*}[!ht]
\centering
\begin{tabular}{lrrrrr}
\toprule
Term & $\beta$ & SE & \textit{df} & $t$ & $p$ \\
\midrule
(Intercept)                            & 4.274    & 0.332 & 14.6  & 12.86    & \textless 0.001 \\
\texttt{cue}$_{\textit{Vision}}$       & $-$0.068 & 0.149 & 322.7 & $-$0.45  & 0.651 \\
\texttt{attractors}                    & $-$0.875 & 0.067 & 322.7 & $-$13.11 & \textless 0.001 \\
\texttt{method}$_{\textit{DiffMean}}$  & $-$1.656 & 0.183 & 322.7 & $-$9.06  & \textless 0.001 \\
\texttt{method}$_{\textit{Probe}}$     & $-$1.675 & 0.183 & 322.7 & $-$9.16  & \textless 0.001 \\
\texttt{model}$_{\textit{Qwen3-VL-4B}}$& 0.460    & 0.160 & 329.4 & 2.88     & 0.004 \\
\bottomrule
\end{tabular}
\caption{Linear mixed-effects model results}
\label{tab:lmm-avgodds}
\end{table*}

\section{Additional Mechanistic Results}
\label{app:more-mech-results}

\subsection{Replication of mechanistic results across multiple random seeds}
\label{app:seed-replicate}

\begin{figure*}[h]
    \centering
    \includegraphics[width=0.8\textwidth]{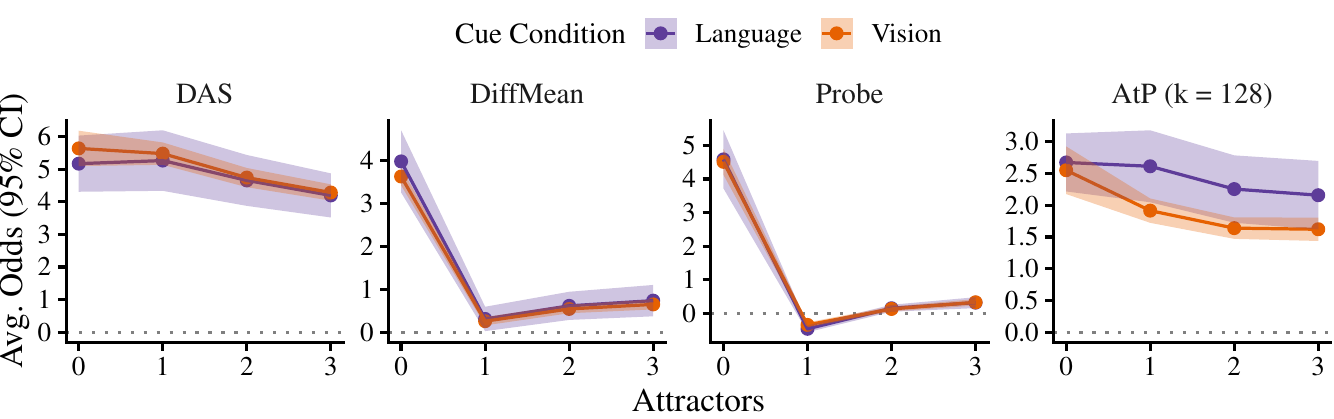}
    \caption{Mechanistic results across four additional
  \texttt{[wug]}/\texttt{[wugs]} embedding pairs
  in \texttt{Qwen-3-VL-2B} across all conditions.}
    \label{fig:lr_sweep_2b_interp}
\end{figure*}

We verify that the generalization of our mechanistic results is not attributable to the random initialization or training dynamics of a particular \texttt{[wug]} and \texttt{[wugs]} embedding, replicating our analyses for four additional embeddings pairs. \Cref{fig:lr_sweep_2b_interp} shows  results on Qwen-3-VL-2B. We find that our results are highly consistent with our main mechanistic findings as described in Section \ref{sec:mechanistic}. One exception is the results on AtP*, where the vision condition lags behind language condition, though only on one or more attractors. Overall, though, the results seem generally consistent.

% \begin{figure}[H]
%     \centering
%     \includegraphics[width=0.8\linewidth]{figures/interp_odds_4b.pdf}
%     \caption{Mechanistic results for LR sweeps ($n=5$ learning rates) in \texttt{Qwen-3-VL-4B} for all conditions.}
%     \label{fig:lr_sweep_4b_interp}
% \end{figure}

\subsection{Selecting top-$k$ activations and activation basis for attribution patching}
\label{app:atp-k-select}
\begin{figure}[h]
    \centering
    \includegraphics[width=0.8\linewidth]{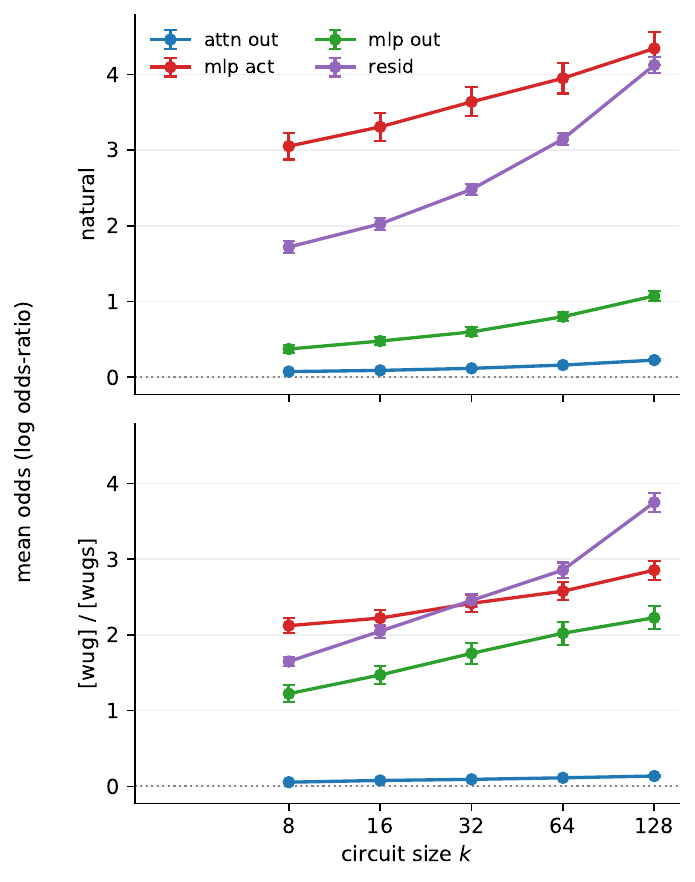}
    \caption{Efficacy of various activation bases and values of top-$k$ activations for causal interventions in \texttt{Qwen-3-VL-2B}.}
    \label{fig:hook_compare}
\end{figure}

It is an empirical question whether a certain activation basis allows for more effective causal interchanges in the attribution patching method, or how the effectiveness of interchanges changes for different numbers of top-$k$ activations. We compare the mean odds of patching top-$k$ activations from attention heads, the residual stream, and MLP activations pre and post-linearity, for $k={2^3...2^7}$ activations in Qwen-3-VL-2B. These results can be seen in Figure \ref{fig:hook_compare}. We observe the greatest effectiveness of causal interventions using the top $k=128$ MLP intermediate activations as an activation basis in natural stimuli, consistent with findings from \citet{jafari2025relp}. Interestingly, we find differing results in generalization to \texttt{[wug]} and \texttt{[wugs]} embeddings.

\section{Implementation and Compute Resources}
\label{app:imp-resources}
All experiments in this paper were conducted on either an \texttt{NVIDIA H-100} GPU or an \texttt{NVIDIA RTX 6000 Ada} GPU. We used \texttt{Adam} as the optimization method for training all embeddings and causal intervention representations (for methods with a supervised learning objective). We sampled from the following range of learning rates in our grid search for optimal LR in both \texttt{Qwen-3-VL-2B} and \texttt{Qwen-3-VL-4B}: 0.0001, 0.0003, 0.0005, 0.0007, 0.0009, 0.001, 0.003, 0.005, 0.0075, 0.01, 0.03, 0.05, 0.075, 0.1. All our code is implemented in python, with DAS being implemented in \texttt{pyvene} \citep{wu-etal-2024-pyvene}, log-probabilities and generation using \texttt{minicons} \citep{misra2022minicons} and \texttt{transformers} libraries \citep{wolf-etal-2020-transformers}.

\section{AI usage disclosure statement}
\label{app:ai-use}
LLMs (specifically, OpenAI \texttt{o3}) were used in this project for the purposes of generating novel image and text stimuli.

\end{document}